\documentclass[runningheads]{llncs}

\usepackage{eccv}

\usepackage{eccvabbrv}

\usepackage{graphicx}
\usepackage{booktabs}
\usepackage{multirow}
\usepackage{tcolorbox}
\tcbuselibrary{breakable}
\usepackage{listings}
\usepackage{needspace}
\usepackage{enumitem}
\usepackage{svg}
\usepackage{subcaption}
\usepackage[table]{xcolor}
\usepackage{placeins}
\usepackage[accsupp]{axessibility}  

\definecolor{jimcolor}{RGB}{83, 158, 198}
\definecolor{blue_munsell}{rgb}{0.36, 0.54, 0.66}
\definecolor{blue-violet}{rgb}{0.54, 0.17, 0.89}
\definecolor{byzantine}{rgb}{0.74, 0.2, 0.64}
\definecolor{caputmortuum}{rgb}{0.35, 0.15, 0.13}
\definecolor{alizarin}{rgb}{0.82, 0.1, 0.26}
\definecolor{light_grey}{rgb}{0.6, 0.6, 0.6}

\newif\ifshowcomments

\showcommentstrue

\usepackage{hyperref}

\usepackage{orcidlink}

\begin{document}

\title{The Cost of Reasoning: Chain-of-Thought Induces Overconfidence in Vision-Language Models}

\titlerunning{The Cost of Reasoning in VLMs}

\author{Robert Welch\inst{1,2} \and
Emir Konuk\inst{1,2} \and
Kevin Smith\inst{1,2}}

\authorrunning{R.~Welch et al.}

\institute{KTH Royal Institute of Technology, Stockholm, Sweden
\and
Science for Life Laboratory, Stockholm, Sweden\\
\email{rwe2@kth.se}
}

\maketitle

\newcommand{\suppref}[1]{the supp. material (Sec.~#1)}

\begin{abstract}
Vision-language models (VLMs) are increasingly deployed in high-stakes settings where reliable uncertainty quantification (UQ) is as important as predictive accuracy. Extended reasoning via chain-of-thought (CoT) prompting or reasoning-trained models has become ubiquitous in modern VLM pipelines, yet its effect on UQ reliability remains poorly understood. Our results show that reasoning tends to degrade the  quality of many uncertainty estimates, even when it improves task accuracy. We identify implicit answer conditioning as the primary mechanism: as reasoning traces converge on a conclusion before the final answer is generated, token probabilities increasingly reflect consistency with the model’s own reasoning trace rather than uncertainty about correctness. In effect, the model becomes overconfident in its answer. In contrast, agreement-based consistency remains robust and often improves under reasoning, making it a practical choice for uncertainty estimation in reasoning-enabled VLMs.

  \keywords{Vision-Language Models \and Chain-of-Thought \and Uncertainty Quantification}
\end{abstract}

\section{Introduction}
\label{sec:intro}


\begin{figure}
    \centering
    \includegraphics[width=1.0\linewidth]{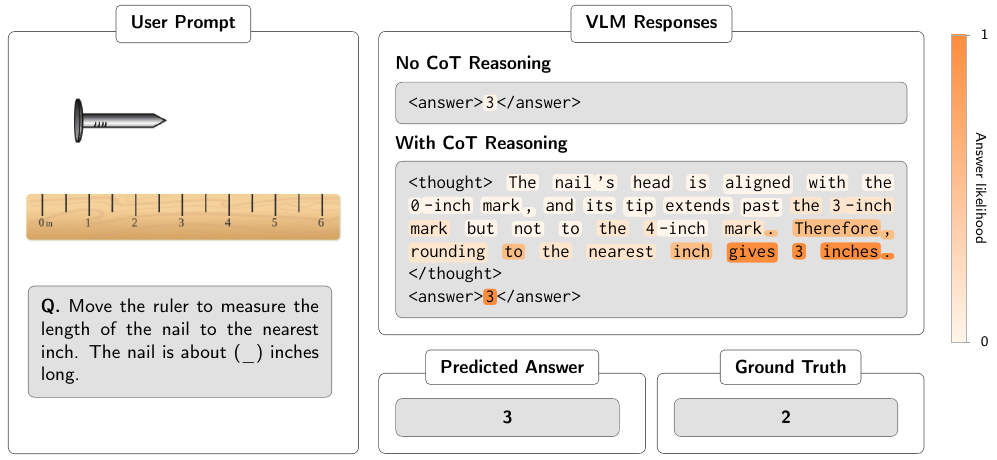}
   \caption{Example of implicit answer conditioning: intermediate reasoning progressively commits the model to its eventual answer before that answer is explicitly generated. Colour indicates the model's confidence in its final predicted answer $\hat{a}$ on the same input, without reasoning ($P(\hat{a}\mid x,\theta)$) and at each reasoning token position $t$ with chain-of-thought reasoning ($P(\hat{a}\mid x, r_{\leq t},\theta)$), where $x$ is the input and $r_{\leq t}$ is the reasoning trace up to token $t$. Although the prediction is incorrect in both cases (3 inches; ground truth: 2), confidence rises sharply as the reasoning trace converges on the wrong answer, exceeding the no-CoT confidence.}

    \label{fig:figure_1}
\end{figure}

Vision-language models (VLMs) are increasingly deployed in high-stakes settings where reliable predictions are as important as accurate ones, from interpreting medical images  \cite{kurz2025benchmarking} to autonomous navigation \cite{xu2024vlm}. Modern VLM pipelines increasingly rely on chain-of-thought (CoT) prompting and reasoning-oriented models that generate intermediate, step-by-step inference before producing a final answer. Reasoning substantially improves performance on challenging multimodal benchmarks requiring multi-step visual and mathematical reasoning \cite{lu2023mathvista,yue2025mmmu}.

However, accuracy alone is insufficient for safe deployment. 
\textit{Selective generation} addresses this issue by ensuring that the models can abstain when uncertain \cite{cole2023selectively}. In this setting, uncertainty quality is determined by how well confidence scores rank correct predictions above incorrect ones, and is typically evaluated using ranking-based metrics such as the Prediction Rejection Ratio (PRR) \cite{malinin2017incorporating,malinin2020uncertainty} and AUGRC \cite{traub2024overcoming}. 

Given the widespread adoption of reasoning, a natural question arises. How does reasoning affect uncertainty estimation in multi-modal generation? One might expect that deliberate reasoning improves both accuracy and UQ reliability. However, across the VLM families and benchmarks we evaluated, we find that this intuition often fails for many uncertainty estimates. While reasoning typically improves task accuracy, it tends to degrade the ranking quality of uncertainty estimates that rely on answer-token likelihoods, including maximum sequence probability, perplexity, mean token entropy, and Monte Carlo sequence entropy \cite{fadeeva2023lm, kuhn2023semantic}. The issue is not merely that additional conditioning changes token probabilities (which is expected) but that reasoning changes what the answer-token likelihood represents. Rather than reflecting uncertainty about correctness, it increasingly captures internal semantic commitment to the model’s own reasoning trace. In multimodal settings, this misalignment is especially consequential: VLMs must ground their conclusions in perceptual evidence that may remain ambiguous even as the reasoning trace converges. Token-level probabilities can signal consistency with the model’s internal reasoning while failing to capture uncertainty about the visual signal. As a result, models may abstain less often on difficult or visually ambiguous samples, producing overconfident errors despite improved average accuracy.

Our analysis suggests that this arises from what we term \textit{implicit answer conditioning}: as a reasoning trace unfolds, it progressively constrains the conditional distribution over the forthcoming answer tokens. By the time the final answer is generated, the model has already committed to a specific conclusion within its own reasoning trace, making the answer highly predictable. This inflates answer-token likelihoods, whether or not the conclusion is correct. Importantly, this effect is not attributable to verbosity: longer reasoning traces are, on average, associated with lower confidence. Rather, it is the conditioning effect of the reasoning context (i.e.~the answer implicit in the reasoning) that concentrates probability mass on the predicted answer. We provide controlled masking interventions and reasoning trace analyses that offer evidence consistent with this explanation. A visual illustration of implicit answer conditioning is shown in \cref{fig:figure_1}.

Importantly, not all uncertainty estimates are equally affected. Those that do not rely directly on answer-token likelihoods, such as agreement-based consistency \cite{wang2022self}, which measures agreement across sampled answers via a majority vote, remain stable and often improve under reasoning. Asking the model to self-report confidence \cite{kadavath2022language, srinivasan2024selective} shows mixed behaviour. 
This paper makes the following contributions:
\begin{enumerate}
\item We empirically demonstrate that reasoning tends to degrade the quality of uncertainty estimates that rely on answer-token likelihood (ATL).
\item We provide evidence pointing towards \textit{implicit answer conditioning} as a key driver of this degradation.
\item We show that applying agreement-based consistency is robust under reasoning, offering a practical solution for UQ in reasoning VLMs.
\end{enumerate}

\section{Related Work}
\label{sec:related_work}

The interaction between chain-of-thought reasoning and uncertainty estimation has recently been explored in text-only large language models. Yoon et al.~\cite{yoon2025reasoning} report that CoT-induced “slow-thinking” behaviours can improve self-reported confidence calibration, while Lyu et al.~\cite{lyu2025calibrating} study whether consistency enhances calibration in LLMs. Lanham et al.~\cite{lanham2023measuring} inspect reasoning traces to analyze their causal role in shaping model predictions. However, these works focus on calibration or reasoning faithfulness in text-only settings. They do not examine how reasoning alters the pragmatic ranking reliability of uncertainty estimates or the selective generation setting.

Crucially, prior studies do not consider multimodal models, where reasoning must reconcile visual perceptual uncertainty with linguistic inference. In multimodal models, the reasoning process can converge on a confident conclusion even if the visual evidence is unclear, producing mismatches between internal confidence and perceptual uncertainty that do not occur in text-only models. To our knowledge, no prior work analyzes how reasoning alters uncertainty ranking in multimodal generation.

Chain-of-thought prompting~\cite{wei2022chain} and reasoning-native (“Thinking”) models~\cite{jaech2024openai, guo2025deepseek} generate intermediate reasoning traces prior to producing final answers, and are now widely adopted. Self-Consistency~\cite{wang2022self} aggregates multiple reasoning paths via majority voting. Despite their widespread use, their impact on uncertainty estimation remains largely unexplored.

\textbf{Uncertainty Estimation in Language and Vision-Language Models.} 
Comprehensive benchmarks such as LM-Polygraph~\cite{fadeeva2023lm} and subsequent large-scale evaluations~\cite{vashurin2025benchmarking, ye2024benchmarking} compare a wide range of white-box and black-box uncertainty methods for LLMs. These studies, however, evaluate static uncertainty quantification for non-reasoning settings and do not analyze how reasoning behaviour systematically modifies them.

In multimodal settings, uncertainty for VLMs has thus far been studied via conformal prediction on multiple-choice tasks without reasoning~\cite{kostumov2024uncertainty}. While useful, conformal methods assume fixed answer sets and do not address open-ended generation or token-level likelihood-based uncertainty. In this paper, we analyze a broad spectrum of uncertainty measures including token-likelihood-based, agreement-based, and introspective measures for multimodal settings.

Beyond benchmark comparisons, prior work explores behavioural proxies for uncertainty. Khan and Fu~\cite{khan2024consistency} examine answer consistency under prompt rephrasing for selective visual question answering. Devic et al.~\cite{devic2025trace} propose reasoning trace length as an uncertainty signal in reasoning-oriented text-only models. While related, these approaches do not analyze how reasoning changes the semantics of token-level likelihood or its reliability as an uncertainty estimate in multimodal models.
Kuhn et al.~\cite{kuhn2023semantic} introduce Semantic Entropy, which estimates predictive uncertainty by sampling multiple generations, clustering them into semantic equivalence classes using an external natural language inference (NLI) model, and computing entropy over clusters. Semantic Entropy captures semantic diversity across sampled outputs, whereas our analysis focuses on how reasoning distorts token-level likelihood within a single conditioned generation. Moreover, Semantic Entropy relies on an auxiliary NLI oracle; we instead evaluate uncertainty signals computed directly from model outputs to enable consistent multimodal comparison.

Kadavath et al.~\cite{kadavath2022language} demonstrate that models can estimate the probability that a given answer is correct via the \emph{P(True)} self-probing signal. Our Self-Reported Confidence measure is closely related; following~\cite{srinivasan2024selective}, we normalize this score and evaluate it under ranking-based selective generation metrics.

\textbf{Evaluating Uncertainty Under Selective Generation.}
We evaluate uncertainty within the selective generation framework, where a model may abstain when its confidence falls below a threshold. This paradigm reflects safety-critical deployment and \textit{assesses uncertainty by how well it ranks correct predictions above incorrect ones under abstention}. Metrics such as the Area Under the Risk–Coverage Curve (AURC)~\cite{geifman2018bias} and the Prediction Rejection Ratio (PRR)~\cite{fadeeva2023lm, malinin2017incorporating} directly quantify this ranking quality. We build on it to show that reasoning systematically alters the reliability of token-level uncertainty signals.

\section{Method}
\label{sec:method}


 In the selective generation framework, the model produces answers for only a subset of inputs (coverage), rejecting uncertain predictions and trading off coverage against error rate (risk). 
Given a confidence score $g(x)$ for input $x$, predictions are accepted only when $g(x) \ge \tau$ for some threshold $\tau$, otherwise we abstain from providing an answer. Varying $\tau$ reveals a trade-off between coverage level and risk, and provides a pragmatic evaluation of how effectively an uncertainty estimate identifies unreliable predictions.

\subsection{Uncertainty Estimation Methods}
\label{sec:UQ_estimates_overview}

Decoder-style VLMs output a probability distribution over a vocabulary $\mathcal{V}$ containing hundreds of thousands of tokens. To decide whether to answer or abstain, this distribution must be reduced to a scalar uncertainty score. 
We evaluate established uncertainty estimation strategies, grouped into three categories: Answer-Token Likelihood (ATL) Estimates, Self-Probing Estimates, and Agree\-ment-Based Estimates.


\subsubsection{Answer-Token Likelihood (ATL) Estimates.}

These measures compress token-level probabilities into a single scalar calculated over the answer tokens. Let $y = (y_1, \dots, y_{L_{\text{ans}}})$ denote the answer sequence, where $y_l$ is the $l$-th token.

\paragraph{Maximum Sequence Probability (MSP).}

The joint probability of the generated answer sequence is used directly:

\begin{equation}
P(y \mid x,\theta)
   = \prod_{l=1}^{L_{\text{ans}}} P(y_l \mid y_{<l}, x,\theta)
\end{equation}

\noindent Because probabilities are multiplicative across tokens, MSP decreases exponentially with sequence length, making it sensitive to answer length.

\paragraph{Perplexity (PPL).}

Perplexity mitigates this length bias by normalizing the log-likelihood by sequence length:

\begin{equation}
\operatorname{PPL}(y \mid x,\theta)
   = \exp\!\Bigl(
          -\frac{1}{L_{\text{ans}}}
           \sum_{l=1}^{L_{\text{ans}}}
           \log P(y_l \mid y_{<l},x,\theta)
        \Bigr)
\end{equation}

\paragraph{Mean Token Entropy (MTE).}

This measure averages the Shannon entropy of the next-token distribution across all answer positions:

\begin{equation}
\overline{H}(y \mid x,\theta)
   = \frac{1}{L_{\text{ans}}}
   \sum_{l=1}^{L_{\text{ans}}}
   \left[
      -\sum_{v \in \mathcal{V}}
         P(v \mid y_{<l},x,\theta)
         \log P(v \mid y_{<l},x,\theta)
   \right]
\end{equation}

\paragraph{Monte Carlo Sequence Entropy (MC-SE).}

Following \cite{kuhn2023semantic}, predictive uncertainty can also be estimated from sampled generations. Given $K$ sampled sequences $y^{(k)} \sim P(\cdot \mid x,\theta)$:

\begin{equation}
H_{\text{MC-SE}}(x;\theta)
= -\frac{1}{K}\sum_{k=1}^{K}\log P(y^{(k)}\mid x,\theta)
\end{equation}

\noindent To reduce bias introduced by variable sequence length, we also evaluate the length-normalized variant:

\begin{equation}
H_{\text{MC-NSE}}(x;\theta)
= -\frac{1}{K}\sum_{k=1}^{K}\frac{1}{L_{\text{ans},k}}\log P(y^{(k)}\mid x,\theta)
\end{equation}

\noindent Importantly, both MC-SE and MC-NSE aggregate token probabilities within each sampled sequence independently. They therefore do not explicitly measure disagreement across generations. Thus, different answers may produce low entropy if each sequence individually receives high token likelihood.

\subsubsection{Self-Probing.}

Rather than aggregating token probabilities, self-probing methods directly query the model for its confidence in its own prediction.

\paragraph{Self-Reported Confidence (SRC).}

Following \cite{srinivasan2024selective}, we elicit confidence through meta-prompting. Given a query $Q$, image $I$, and model answer $A$, we ask:
“\texttt{Is the above answer correct? Answer `yes' or `no'.}”
Let $\mathcal{X} = \{Q, I, A\}$ denote the full context. The confidence score is defined as the normalized probability of the \texttt{yes} token:

\begin{equation}
P_{\text{src}}
=
\frac{P(\text{yes} \mid \mathcal{X}, \theta)}
     {P(\text{yes} \mid \mathcal{X}, \theta) +
      P(\text{no} \mid \mathcal{X}, \theta)}
\end{equation}

\subsubsection{Agreement-Based Estimates.}

Agreement-based methods estimate uncertainty by measuring variation across multiple stochastic generations.

\paragraph{Consistency.}

Following \cite{wang2022self}, we sample $K$ answers and measure the fraction agreeing with the majority vote. Let $a^{(k)}$ denote the parsed answer from the $k$-th generation and $A^\ast$ the majority answer:

\begin{equation}
\text{Consistency}
=
\frac{1}{K}
\sum_{k=1}^{K}
\mathbf{1}[a^{(k)} = A^\ast]
\end{equation}

\noindent Unlike the previous methods, consistency is a black-box uncertainty estimate: it does not require access to the model’s token probabilities. By capturing disagreement between answers it can detect uncertainty arising from multiple plausible answers which token-likelihood estimates may not reflect if they become overly confident.

\subsection{Evaluating Uncertainty Quality}
\label{sec:selective_generation_metrics}

In selective generation, a model abstains whenever its confidence falls below a threshold. 
The effectiveness of this decision rule therefore \textit{depends on how well the uncertainty score 
orders correct predictions ahead of incorrect ones}. Metrics that isolate \textit{ranking quality} 
thus provide the clearest view of uncertainty reliability. 
Traditional calibration metrics such as Expected Calibration Error (ECE) are less informative 
in this setting because many effective uncertainty estimates are not calibrated probabilities~\cite{ren2023self} and well-calibrated confidence scores do not necessarily produce reliable rankings \cite{jaeger2022call}. For completeness, we report ECE results for applicable uncertainty estimates in \suppref{A}.

\subsubsection{Prediction Rejection Ratio (PRR).}

Following \cite{fadeeva2023lm}, our primary metric is the Prediction Rejection Ratio (PRR) \cite{malinin2017incorporating, malinin2020uncertainty}. Many selective generation metrics couple uncertainty estimates with the model’s accuracy, making it difficult to determine whether performance changes arise from improved predictions or better uncertainty ranking.
PRR isolates the quality of the uncertainty ranking by comparing it to a random baseline and an oracle ranking. It is defined as:

\begin{equation}
PRR = \frac{AR_{\text{unc}}}{AR_{\text{orc}}}
\end{equation}

\noindent where $AR_{\text{unc}}$ is the area between the rejection curve produced by the uncertainty estimate and the random baseline, and $AR_{\text{orc}}$ is the area between the oracle rejection curve (representing a perfect ranking) and the random baseline.
The rejection curve plots the model’s error rate as increasingly uncertain samples are rejected. PRR therefore provides a measure that is largely invariant to overall model accuracy, allowing us to isolate how reasoning affects the ranking reliability of uncertainty estimates.

\subsubsection{Spearman correlation.} We also report the Spearman rank correlation between each uncertainty score and sample correctness. Spearman correlation provides an additional nonparametric measure of ranking agreement between uncertainty scores and correctness.
We invert uncertainty scores where necessary so that higher values indicate higher confidence.

\subsubsection{Risk-Coverage and AUGRC.}

As a secondary metric reflecting practical deployment performance, we analyze the Risk-Coverage (RC) curve. Coverage denotes the fraction of predictions the model accepts, while risk corresponds to the error rate among those accepted predictions.
The Area Under the Risk-Coverage Curve (AURC) \cite{geifman2018bias} is widely used but was originally developed for classification and can exhibit high variance at low coverage levels. Following \cite{traub2024overcoming}, we therefore report the Area Under the Generalized Risk-Coverage Curve (AUGRC).
The generalized risk measures the \textit{probability that a prediction is both accepted and incorrect}. For model $m$, confidence scoring function $g$, and error function $l$, the generalized risk at threshold $\tau$ is defined as:

\begin{equation}
\mathcal{R}_{gen}(m,g)(\tau)
=
\frac{1}{N}
\sum_{i=1}^{N}
l(m(x_i), y_i)
\cdot
\mathbb{I}[g(x_i) \ge \tau]
\end{equation}

\noindent Integrating this quantity across all coverage levels yields AUGRC.
Unlike PRR, which isolates ranking quality, AUGRC reflects the overall behaviour of the deployed system by jointly capturing predictive accuracy and the effectiveness of uncertainty estimates for abstention.

\section{Experimental Setup}
\label{sec:experiments}

\subsubsection{Tasks and Datasets.}

We evaluate on benchmarks spanning visual classification and reasoning-intensive multimodal QA to analyze how uncertainty estimates behave under varying reasoning demands.

We evaluate on four datasets spanning tasks with varying reasoning requirements: 
\textbf{OK-VQA}~\cite{okvqa}, a knowledge-intensive visual question answering benchmark; 
\textbf{MathVista}~\cite{lu2023mathvista}, which requires multi-step visual mathematical reasoning; 
\textbf{MMMU-Pro-Vision}~\cite{yue2025mmmu}, the vision-only subset of an expert-level multimodal reasoning benchmark; 
and \textbf{Oxford-IIIT Pet}~\cite{parkhi2012cats}, a fine-grained image classification dataset. 
Oxford-IIIT Pet serves as a control task (pure visual classification) where reasoning provides little benefit, allowing us to test whether changes in uncertainty estimates arise independently of reasoning-driven accuracy gains.

\subsubsection{Models.}
We evaluate recent open-weight vision-language models with strong reasoning capabilities: \textbf{Gemma3-4B-IT}~\cite{gemmateam2025gemma3technicalreport}, \textbf{InternVL3-9B} \cite{Zhu2025InternVL3EA} and \textbf{Qwen3-VL} (8B, 32B) \cite{bai2025qwen3}. To isolate inference-time reasoning effects, we compare instruct\-ion-tuned (-IT) models with and without chain-of-thought (CoT) prompting while keeping the underlying model fixed. We additionally compare \textbf{Qwen3-VL-8B-IT} (without CoT prompting) with \textbf{Qwen3-VL-8B-Thinking}, a reasoning-trained variant of the same base model, to assess whether the observed effects persist when reasoning is intrinsic to the model rather than induced through prompting.

\subsubsection{Reasoning Protocol.}
Our CoT prompting follows the standard zero-shot reasoning paradigm~\cite{kojima2022large}, encouraging the model to generate intermediate reasoning before producing a final answer. For MathVista and OK-VQA we include dataset-specific instructions, and for OK-VQA we additionally provide four text-only in-context examples following common evaluation practice. Full prompt templates are provided in \suppref{B}. For the comparison involving the reasoning-trained model (Qwen3-VL-8B-Thinking), we use the same prompts without CoT instructions to ensure that differences arise from the model rather than prompt design.

\subsubsection{Inference and Decoding.}
All models are evaluated using their recommended decoding parameters and are provided in \suppref{C} unless otherwise specified. For uncertainty estimates requiring multiple generations we generate $K=10$ stochastic outputs per input using the same decoding configuration as standard inference. An ablation over different values of $K$ is provided in \suppref{D}. Final answers are extracted using the standardized format described in \suppref{B and E}, ensuring that uncertainty estimates are computed over the final answer span rather than the intermediate reasoning trace.

\section{Main Results}
\label{sec:Main_Results}

We report results for Qwen3-VL-8B-IT with and without CoT prompting in \cref{tab:Main_Results_qwen3-vl-8b-it-all_datasets}. We then compare Qwen3-VL-8B-IT (no CoT prompting) against Qwen3-VL-8B-Thinking in \cref{tab:thinking_all_models}. Results for Gemma3-4B-IT, InternVL3-9B and Qwen3-VL-32B-IT are provided in \suppref{F} and exhibit the same trends.
Across settings, we first observe that ATL-based estimates perform well in the absence of reasoning. However, once models generate reasoning traces, their uncertainty ranking quality deteriorates even when their accuracy improves, while agreement-based uncertainty estimates remain reliable.

\subsubsection{CoT Reasoning Degrades Uncertainty Estimates.}
\label{sec:cot_results}
In \cref{tab:Main_Results_qwen3-vl-8b-it-all_datasets}, we observe that without CoT prompting, ATL-based estimates often achieve the strongest performance across AUGRC, PRR, and Spearman. Under CoT prompting, however, their ranking quality, as measured by PRR and Spearman, deteriorates substantially. Consistency, however, remains reliable across all datasets and settings. The only exception is CoT on Oxford-IIIT Pet, where self-reported confidence (SRC) achieves comparable performance to consistency.

We further note that for the ATL-based estimates, AUGRC improves on reasoning intensive datasets with CoT (MathVista and MMMU-Pro-Vision). However, these improvements occur only when CoT yields substantial gains in predictive accuracy. As explained in \cref{sec:selective_generation_metrics}, AUGRC is influenced by both the predictive accuracy of the model and ranking quality of the UQ estimates. Both PRR and Spearman degrade indicating worsening ranking quality. We can attribute the lower AUGRC under CoT reasoning to improved task performance.

\begin{table}[tb]
\centering
\small
\setlength{\tabcolsep}{6pt}
\renewcommand{\arraystretch}{1.05}
\caption{Selective generation results for Qwen3-VL-8B-IT with and without CoT prompting. Accuracy is reported as (Acc: \textit{No CoT / CoT}) for each setting.}
\label{tab:Main_Results_qwen3-vl-8b-it-all_datasets}
\resizebox{\linewidth}{!}{%
\begin{tabular}{ll ccc ccc ccc}
\toprule
 & & \multicolumn{3}{c}{\textbf{No CoT}} & \multicolumn{3}{c}{\textbf{CoT}} & \multicolumn{3}{c}{\textbf{$\Delta$ (CoT$-$ No CoT)}} \\
\cmidrule(lr){3-5} \cmidrule(lr){6-8} \cmidrule(lr){9-11}
\textbf{Type} & \textbf{UQ Estimate} & \small{AUGRC$\downarrow$} & \small{PRR$\uparrow$} & \small{Spear.$\uparrow$} & \small{AUGRC$\downarrow$} & \small{PRR$\uparrow$} & \small{Spear.$\uparrow$} & \small{AUGRC} & \small{PRR} & \small{Spear.} \\
\midrule
\rowcolor{gray!5} \multicolumn{8}{l}{\textbf{Oxford-IIIT Pet}} \\
\multirow{4}{*}{\shortstack[l]{\textit{Single-Generation} \\ \scriptsize (Acc: 79.3 / 76.2)}} & MSP & 0.056 & 0.710 & 0.404 & 0.110 & 0.231 & 0.075 & 0.053 & -0.478 & -0.329 \\
 & PPL & 0.057 & 0.711 & 0.399 & 0.111 & 0.211 & 0.061 & 0.054 & -0.500 & -0.338 \\
 & MTE & \textbf{0.053} & \textbf{0.735} & \textbf{0.431} & 0.112 & 0.205 & 0.056 & 0.059 & -0.530 & -0.376 \\
 & SRC & 0.060 & 0.709 & 0.377 & \textbf{0.100} & \textbf{0.314} & \textbf{0.151} & 0.041 & -0.394 & -0.226 \\
\addlinespace[0.3em] \cmidrule{2-11} \addlinespace[0.3em]
\multirow{3}{*}{\shortstack[l]{\textit{Multi-Generation} \\ \scriptsize (Acc: 79.2 / 76.8)}} & MC-SE & 0.062 & 0.645 & 0.360 & 0.109 & 0.185 & 0.055 & 0.048 & -0.460 & -0.305 \\
 & MC-NSE & 0.064 & 0.625 & 0.337 & 0.112 & 0.147 & 0.030 & 0.048 & -0.478 & -0.307 \\
 & Consistency & 0.081 & 0.525 & 0.284 & \textbf{0.100} & 0.264 & 0.146 & 0.019 & -0.262 & -0.138 \\
\midrule \rowcolor{gray!5} \multicolumn{8}{l}{\textbf{OK-VQA }} \\
\multirow{4}{*}{\shortstack[l]{\textit{Single-Generation} \\ \scriptsize (Acc: 55.9 / 52.1)}} & MSP & \textbf{0.167} & \textbf{0.525} & 0.402 & 0.208 & 0.241 & 0.231 & 0.041 & -0.284 & -0.171 \\
 & PPL & 0.170 & 0.498 & 0.374 & 0.216 & 0.176 & 0.170 & 0.046 & -0.322 & -0.203 \\
 & MTE & 0.168 & 0.510 & 0.392 & 0.216 & 0.174 & 0.172 & 0.048 & -0.337 & -0.220 \\
 & SRC & 0.175 & 0.465 & 0.341 & 0.194 & 0.441 & 0.339 & 0.019 & -0.024 & -0.002 \\
\addlinespace[0.3em] \cmidrule{2-11} \addlinespace[0.3em]
\multirow{3}{*}{\shortstack[l]{\textit{Multi-Generation} \\ \scriptsize (Acc: 56.0 / 54.1)}} & MC-SE & 0.167 & 0.515 & 0.395 & 0.196 & 0.322 & 0.244 & 0.030 & -0.192 & -0.151 \\
 & MC-NSE & 0.171 & 0.488 & 0.366 & 0.202 & 0.274 & 0.202 & 0.031 & -0.214 & -0.164 \\
 & Consistency & \textbf{0.167} & 0.436 & \textbf{0.422} & \textbf{0.171} & \textbf{0.478} & \textbf{0.452} & 0.004 & 0.043 & 0.030 \\
\midrule \rowcolor{gray!5} \multicolumn{8}{l}{\textbf{MMMU-Pro-Vision}} \\
\multirow{4}{*}{\shortstack[l]{\textit{Single-Generation} \\ \scriptsize (Acc: 38.1 / 46.0)}} & MSP & 0.264 & 0.450 & 0.326 & 0.256 & 0.123 & 0.165 & -0.008 & -0.326 & -0.161 \\
 & PPL & 0.264 & 0.450 & 0.326 & 0.256 & 0.123 & 0.165 & -0.008 & -0.326 & -0.161 \\
 & MTE & \textbf{0.263} & \textbf{0.479} & \textbf{0.331} & 0.234 & 0.376 & 0.251 & -0.029 & -0.103 & -0.081 \\
 & SRC & 0.276 & 0.380 & 0.240 & 0.234 & 0.409 & 0.255 & -0.042 & 0.029 & 0.014 \\
\addlinespace[0.3em] \cmidrule{2-11} \addlinespace[0.3em]
\multirow{3}{*}{\shortstack[l]{\textit{Multi-Generation} \\ \scriptsize (Acc: 38.5 / 48.0)}} & MC-SE & \textbf{0.263} & 0.447 & 0.323 & 0.233 & 0.315 & 0.204 & -0.030 & -0.132 & -0.120 \\
 & MC-NSE & \textbf{0.263} & 0.447 & 0.323 & 0.233 & 0.316 & 0.204 & -0.030 & -0.132 & -0.119 \\
 & Consistency & 0.285 & 0.107 & 0.189 & \textbf{0.215} & \textbf{0.434} & \textbf{0.340} & -0.070 & 0.327 & 0.150 \\
\midrule \rowcolor{gray!5} \multicolumn{8}{l}{\textbf{MathVista}} \\
\multirow{4}{*}{\shortstack[l]{\textit{Single-Generation} \\ \scriptsize (Acc: 63.6 / 70.4)}} & MSP & 0.143 & 0.432 & 0.284 & 0.126 & 0.219 & 0.170 & -0.017 & -0.213 & -0.114 \\
 & PPL & 0.148 & 0.406 & 0.243 & 0.132 & 0.178 & 0.122 & -0.017 & -0.228 & -0.121 \\
 & MTE & 0.146 & 0.413 & 0.262 & 0.130 & 0.171 & 0.138 & -0.016 & -0.241 & -0.124 \\
 & SRC & 0.143 & \textbf{0.473} & 0.286 & 0.094 & 0.557 & 0.411 & -0.048 & 0.084 & 0.125 \\
\addlinespace[0.3em] \cmidrule{2-11} \addlinespace[0.3em]
\multirow{3}{*}{\shortstack[l]{\textit{Multi-Generation} \\ \scriptsize (Acc: 64.1 / 73.6)}} & MC-SE & \textbf{0.139} & 0.440 & \textbf{0.292} & 0.109 & 0.279 & 0.182 & -0.030 & -0.161 & -0.111 \\
 & MC-NSE & 0.148 & 0.395 & 0.228 & 0.119 & 0.181 & 0.104 & -0.029 & -0.215 & -0.124 \\
 & Consistency & 0.148 & 0.345 & 0.261 & \textbf{0.068} & \textbf{0.684} & \textbf{0.542} & -0.081 & 0.339 & 0.280 \\
\bottomrule
\end{tabular}
}
\end{table}

\subsubsection{Reasoning-Trained Models Exhibit the Same Pattern}
\label{sec:thinking_results}
We compare a reasoning-trained model (Qwen3-VL-8B-Thinking), which produces reasoning traces without prompting, with its non-reasoning counterpart (Qwen3-VL-8B-IT) in \cref{tab:thinking_all_models}. The same degradation pattern observed with CoT prompting reappears and is amplified, indicating that the effect is not specific to prompting but instead reflects a more general property of reasoning in these models.

Across datasets, we observe the same trend as in \cref{tab:Main_Results_qwen3-vl-8b-it-all_datasets}. ATL-based estimates perform strongly for the Instruct model that does not reason, but ranking quality (PRR and Spearman) deteriorates substantially for the Thinking model. 
The degradation of ATL-based estimates is even more pronounced for the Thinking model than for the CoT-prompted Instruct model. For example, MTE PRR drops from $0.735$ to $0.086$ on Oxford-IIIT Pet and from $0.413$ to $-0.056$ on MathVista. Again, Consistency improves under reasoning, achieving the strongest average performance. Improvements in AUGRC for reasoning coincide with substantial gains in predictive accuracy, indicating that lower AUGRC again reflects improved task performance rather than improved uncertainty ranking.

\begin{table}[tb]
\centering
\small
\setlength{\tabcolsep}{6pt}
\renewcommand{\arraystretch}{1.2}
\caption{Selective generation results for Qwen3-VL-8B-IT (Instruct; no CoT) and Qwen3-VL-8B-Thinking (intrinsic reasoning, no CoT prompting). Accuracy is reported as \textit{(Acc: Instruct / Thinking)} for each setting.}
\label{tab:thinking_all_models}
\resizebox{\linewidth}{!}{%
\begin{tabular}{ll ccc ccc ccc}
\toprule
 & & \multicolumn{3}{c}{\textbf{Instruct}} & \multicolumn{3}{c}{\textbf{Thinking}} & \multicolumn{3}{c}{\textbf{$\Delta$ (Thinking$-$Instruct)}} \\
\cmidrule(lr){3-5} \cmidrule(lr){6-8} \cmidrule(lr){9-11}
\textbf{Type} & \textbf{UQ Estimate} & \small{AUGRC$\downarrow$} & \small{PRR$\uparrow$} & \small{Spear.$\uparrow$} & \small{AUGRC$\downarrow$} & \small{PRR$\uparrow$} & \small{Spear.$\uparrow$} & \small{AUGRC} & \small{PRR} & \small{Spear.} \\
\midrule
\rowcolor{gray!5} \multicolumn{8}{l}{\textbf{Oxford-IIIT Pet}} \\
\multirow{4}{*}{\shortstack[l]{\textit{Single-Generation} \\ \scriptsize (Acc: 79.3 / 77.2)}} & MSP & 0.056 & 0.710 & 0.404 & 0.110 & 0.120 & 0.032 & 0.054 & -0.590 & -0.373 \\
 & PPL & 0.057 & 0.711 & 0.399 & 0.112 & 0.089 & 0.021 & 0.055 & -0.621 & -0.378 \\
 & MTE & \textbf{0.053} & \textbf{0.735} & \textbf{0.431} & 0.112 & 0.086 & 0.019 & 0.059 & -0.649 & -0.412 \\
 & SRC & 0.060 & 0.709 & 0.377 & \textbf{0.075} & \textbf{0.538} & \textbf{0.320} & 0.016 & -0.171 & -0.057 \\
\addlinespace[0.3em] \cmidrule{2-11} \addlinespace[0.3em]
\multirow{3}{*}{\shortstack[l]{\textit{Multi-Generation} \\ \scriptsize (Acc: 79.2 / 78.1)}} & MC-SE & 0.062 & 0.645 & 0.360 & 0.121 & -0.057 & -0.096 & 0.059 & -0.702 & -0.455 \\
 & MC-NSE & 0.064 & 0.625 & 0.337 & 0.123 & -0.089 & -0.111 & 0.059 & -0.714 & -0.447 \\
 & Consistency & 0.081 & 0.525 & 0.284 & 0.096 & 0.223 & 0.119 & 0.015 & -0.302 & -0.165 \\
\midrule \rowcolor{gray!5} \multicolumn{8}{l}{\textbf{OK-VQA }} \\
\multirow{4}{*}{\shortstack[l]{\textit{Single-Generation} \\ \scriptsize (Acc: 55.9 / 56.0)}} & MSP & \textbf{0.167} & \textbf{0.525} & 0.402 & 0.207 & 0.110 & 0.097 & 0.040 & -0.415 & -0.305 \\
 & PPL & 0.170 & 0.498 & 0.374 & 0.211 & 0.071 & 0.064 & 0.040 & -0.428 & -0.309 \\
 & MTE & 0.168 & 0.510 & 0.392 & 0.212 & 0.065 & 0.055 & 0.044 & -0.446 & -0.337 \\
 & SRC & 0.175 & 0.465 & 0.341 & 0.190 & 0.190 & 0.258 & 0.015 & -0.276 & -0.083 \\
\addlinespace[0.3em] \cmidrule{2-11} \addlinespace[0.3em]
\multirow{3}{*}{\shortstack[l]{\textit{Multi-Generation} \\ \scriptsize (Acc: 56.0 / 58.2)}} & MC-SE & 0.167 & 0.515 & 0.395 & 0.180 & 0.278 & 0.215 & 0.013 & -0.236 & -0.180 \\
 & MC-NSE & 0.171 & 0.488 & 0.366 & 0.185 & 0.239 & 0.177 & 0.014 & -0.249 & -0.189 \\
 & Consistency & \textbf{0.167} & 0.436 & \textbf{0.422} & \textbf{0.151} & \textbf{0.502} & \textbf{0.457} & -0.016 & 0.066 & 0.035 \\
\midrule \rowcolor{gray!5} \multicolumn{8}{l}{\textbf{MMMU-Pro-Vision}} \\
\multirow{4}{*}{\shortstack[l]{\textit{Single-Generation} \\ \scriptsize (Acc: 38.1 / 56.6)}} & MSP & 0.264 & 0.450 & 0.326 & 0.174 & 0.357 & 0.305 & -0.089 & -0.093 & -0.021 \\
 & PPL & 0.264 & 0.450 & 0.326 & 0.174 & 0.357 & 0.305 & -0.090 & -0.092 & -0.021 \\
 & MTE & \textbf{0.263} & \textbf{0.479} & \textbf{0.331} & 0.173 & 0.403 & 0.310 & -0.090 & -0.077 & -0.022 \\
 & SRC & 0.276 & 0.380 & 0.240 & 0.210 & 0.112 & 0.050 & -0.066 & -0.268 & -0.191 \\
\addlinespace[0.3em] \cmidrule{2-11} \addlinespace[0.3em]
\multirow{3}{*}{\shortstack[l]{\textit{Multi-Generation} \\ \scriptsize (Acc: 38.5 / 58.5)}} & MC-SE & \textbf{0.263} & 0.447 & 0.323 & 0.152 & 0.556 & 0.390 & -0.110 & 0.109 & 0.067 \\
 & MC-NSE & \textbf{0.263} & 0.447 & 0.323 & 0.152 & \textbf{0.557} & 0.390 & -0.110 & 0.109 & 0.067 \\
 & Consistency & 0.285 & 0.107 & 0.189 & \textbf{0.149} & 0.455 & \textbf{0.453} & -0.136 & 0.348 & 0.264 \\
\midrule \rowcolor{gray!5} \multicolumn{8}{l}{\textbf{MathVista}} \\
\multirow{4}{*}{\shortstack[l]{\textit{Single-Generation} \\ \scriptsize (Acc: 63.6 / 80.3)}} & MSP & 0.143 & 0.432 & 0.284 & 0.104 & 0.011 & -0.049 & -0.038 & -0.421 & -0.333 \\
 & PPL & 0.148 & 0.406 & 0.243 & 0.108 & -0.050 & -0.078 & -0.041 & -0.456 & -0.321 \\
 & MTE & 0.146 & 0.413 & 0.262 & 0.108 & -0.056 & -0.084 & -0.038 & -0.469 & -0.346 \\
 & SRC & 0.143 & \textbf{0.473} & 0.286 & 0.079 & 0.291 & 0.174 & -0.064 & -0.182 & -0.112 \\
\addlinespace[0.3em] \cmidrule{2-11} \addlinespace[0.3em]
\multirow{3}{*}{\shortstack[l]{\textit{Multi-Generation} \\ \scriptsize (Acc: 64.1 / 80.7)}} & MC-SE & \textbf{0.139} & \textbf{0.440} & \textbf{0.292} & 0.097 & 0.036 & -0.005 & -0.042 & -0.403 & -0.298 \\
 & MC-NSE & 0.148 & 0.395 & 0.228 & 0.101 & -0.045 & -0.037 & -0.047 & -0.441 & -0.265 \\
 & Consistency & 0.148 & 0.345 & 0.261 & \textbf{0.038} & \textbf{0.767} & \textbf{0.634} & -0.111 & 0.422 & 0.373 \\
\bottomrule
\end{tabular}
}
\end{table}

\subsubsection{Reasoning Induces Systematic Confidence Inflation.}
\begin{figure}[tb]
    \centering
    \includegraphics[width=1.0\linewidth]{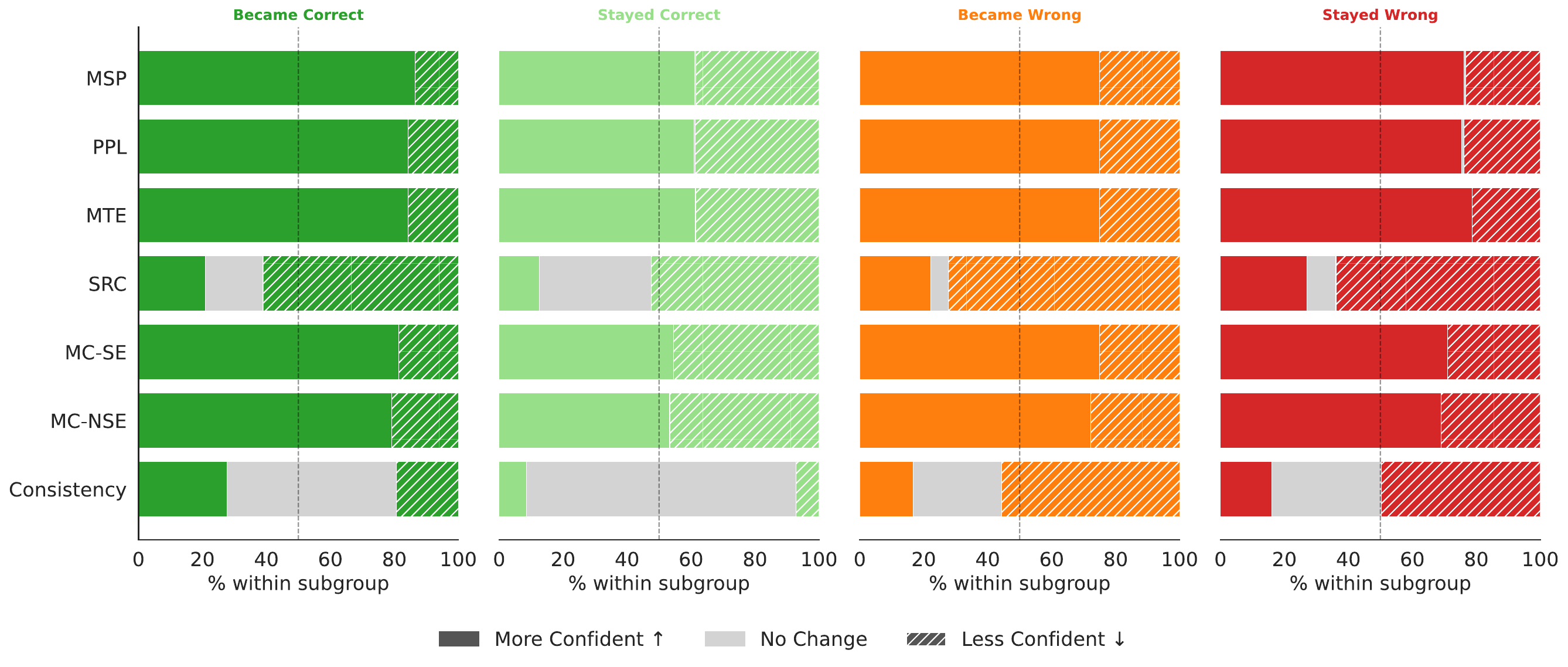}
    \caption{\textbf{Confidence shifts induced by CoT reasoning.} Bars show the percentage of samples whose confidence increases (solid colour), decreases (striped colour), or remains unchanged (gray) when moving from no-CoT to CoT inference, grouped by correctness. ATL-based estimates systematically increase confidence, including samples that become or remain incorrect, indicating confidence inflation unrelated to correctness. Self-reported confidence (SRC) and Consistency do not exhibit systematic inflation.}
    \label{fig:UQ_shift_MathVista}
\end{figure}

To analyze how UQ estimates change under reasoning, we examine
sample-level confidence shifts between no-CoT and CoT inference on MathVista
(\cref{fig:UQ_shift_MathVista}; additional datasets in \suppref{G}). 
We partition samples based on how their correctness changes when moving from
no-CoT to CoT inference. Each sample falls into one of four
groups:
\emph{becomes correct} ($0 \rightarrow 1$),
\emph{remains correct} ($1 \rightarrow 1$),
\emph{becomes incorrect} ($1 \rightarrow 0$), and
\emph{remains incorrect} ($0 \rightarrow 0$).
For each group, we compute the proportion of samples whose 
confidence
increases or decreases under CoT prompting relative to no-CoT prompting.
Across all datasets, ATL-based estimates increase their confidence when using CoT prompting, regardless of their correctness. In contrast, SRC and Consistency do not exhibit systematic confidence inflation.

\section{Analysis}
\label{sec:analysis}

In this section we investigate the mechanism behind the degradation of ATL-based uncertainty estimates under reasoning.

\subsubsection{Reasoning Length does not Drive Confidence Inflation.}
To assess whether the length of the reasoning trace explains the degradation of ATL-based estimates, we compute the partial Spearman correlation between the number of generated reasoning tokens (summary reported in \suppref{H}) and the UQ estimates, while controlling for prediction correctness across all instruction-tuned models. 
As shown in \cref{fig:reasoning_len_UQ_corr}, reasoning length exhibits a negative correlation with all UQ estimates: longer traces correspond to lower confidence. This aligns with prior work linking longer reasoning to greater task difficulty and reduced model confidence \cite{devic2025trace}. This trend contradicts the confidence inflation effect we observe. If reasoning length was the primary driver, longer traces would increase confidence. 

\subsubsection{Implicit Answer Conditioning Drives Confidence Inflation.}
We argue that confidence inflation arises not from reasoning length, but from \textit{implicit answer conditioning}: reasoning traces progressively converge toward a conclusion, increasing the conditional probability of the final answer tokens before they are generated. As the reasoning trace increasingly commits to a particular conclusion, the predicted answer becomes highly predictable given the preceding context. At the token level, this predictability is indistinguishable from genuine model confidence. Consequently, elevated answer-token probabilities reflect commitment to the model’s own reasoning trace rather than reduced uncertainty about correctness.

Formally, let 
$s(\hat{y} \mid x, r)$ denote the ATL confidence score for predicted answer 
$\hat{y}$ conditioned on input $x$ and reasoning trace $r$, and let 
$s(\hat{y} \mid x)$ denote the equivalent score without reasoning. We define 
the reasoning-induced confidence shift as $\Delta s = s(\hat{y} \mid x, r) - 
s(\hat{y} \mid x)$. Reasoning inflates ATL confidence on average, $\mathbb{E}[\Delta s] > 0$. The signature of implicit answer conditioning is that this inflation is \emph{correctness-agnostic}: the model becomes more confident in its predicted answer whether or not that answer is correct,

\begin{equation}
    \mathbb{E}[\Delta s \mid \hat{y}=y^*] \approx \mathbb{E}[\Delta s \mid \hat{y}\neq y^*]
\end{equation}

Because the added confidence is not tied to correctness, it does not help separate correct from incorrect predictions, leaving ATL a less faithful indicator of correctness. This is consistent with the ranking degradation reported in \cref{tab:Main_Results_qwen3-vl-8b-it-all_datasets,tab:thinking_all_models}. Additional evidence is provided in \suppref{I}: the final-answer MSP rises at a near-identical rate for correct and incorrect predictions as the reasoning trace progresses, consistent with a correctness-agnostic confidence shift rather than one driven by genuine gains in predictive accuracy.

To further test whether implicit answer conditioning occurs under standard CoT prompting, we examine the relationship between answer mentions in the reasoning trace and confidence. We use answer frequency as a proxy for conditioning strength: repeated mentions indicate stronger contextual commitment, though this proxy captures only the most explicit form of the effect.

\begin{figure}[t]
    \centering
    \begin{subfigure}[t]{0.24\textwidth}
        \centering
        \includegraphics[width=\textwidth]{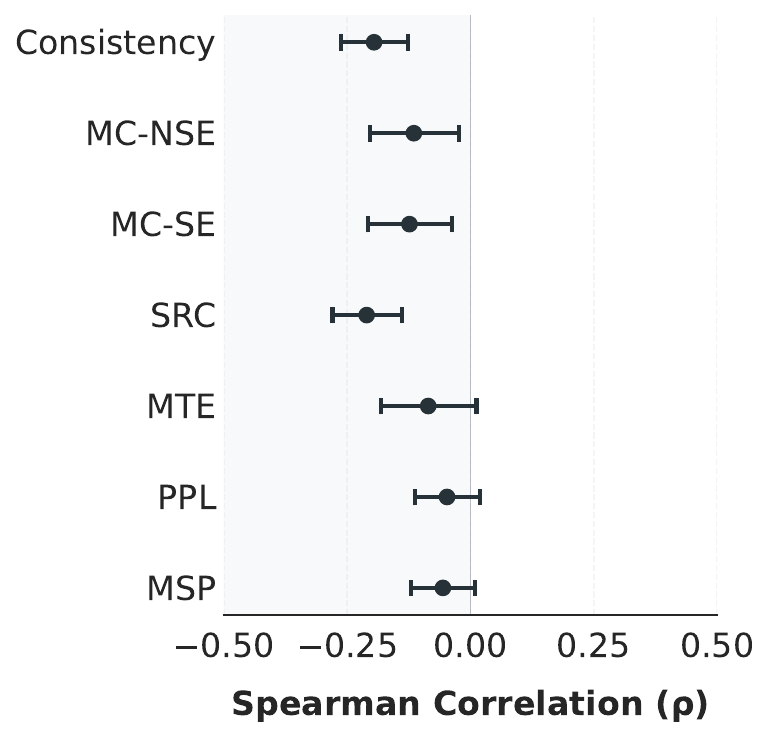}
        \caption{Reasoning Length vs Confidence}
        \label{fig:reasoning_len_UQ_corr}
    \end{subfigure}%
    \hfill
    \begin{subfigure}[t]{0.24\textwidth}
        \centering
        \includegraphics[width=\textwidth]{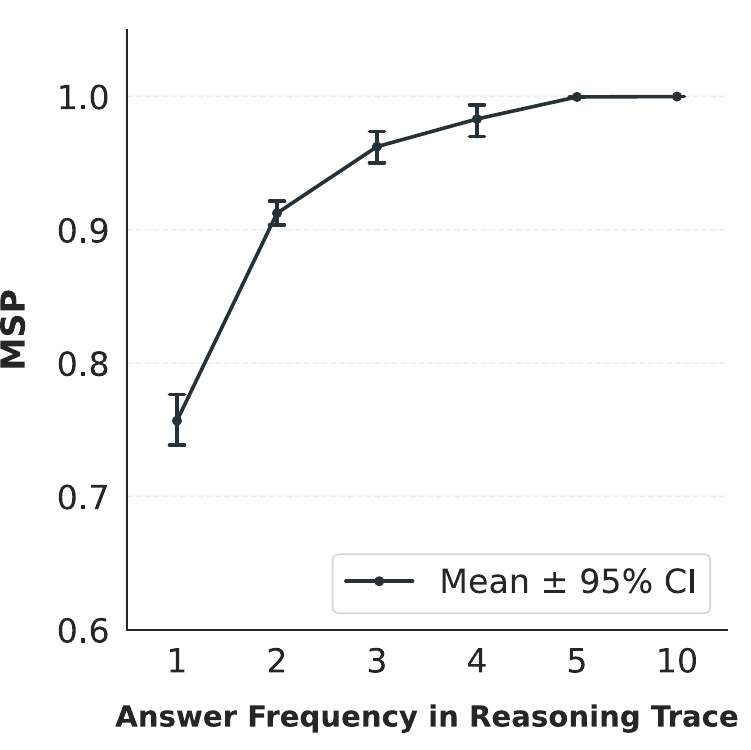}
        \caption{Confidence vs Answer Frequency}
        \label{fig:answer_freq_32b}
    \end{subfigure}%
    \hfill
    \begin{subfigure}[t]{0.24\textwidth}
        \centering
        \includegraphics[width=\textwidth]{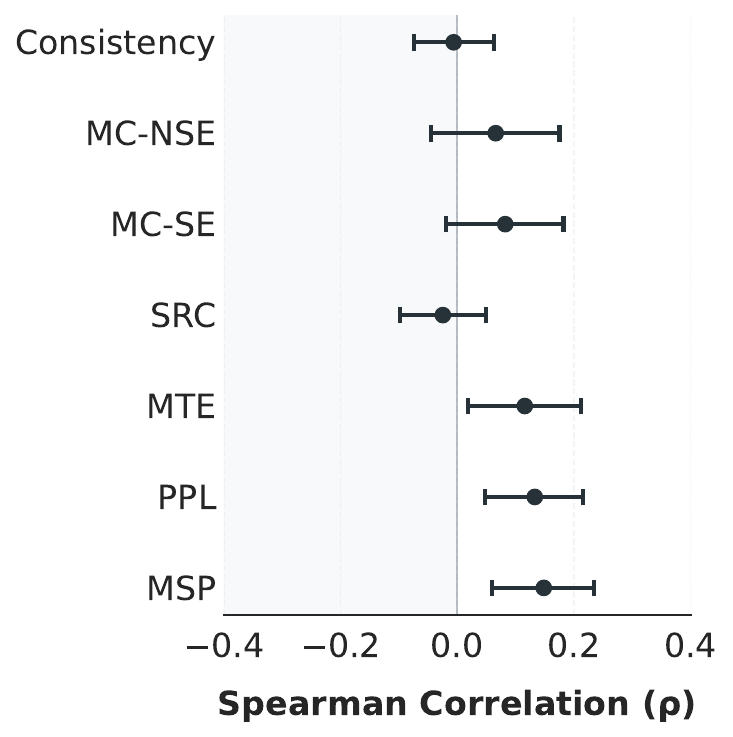}
        \caption{Answer Frequency \& Confidence Correlation \\ (all samples)}
        \label{fig:answer_freq_corr_all}
    \end{subfigure}%
    \hfill
    \begin{subfigure}[t]{0.24\textwidth}
        \centering
        \includegraphics[width=\textwidth]{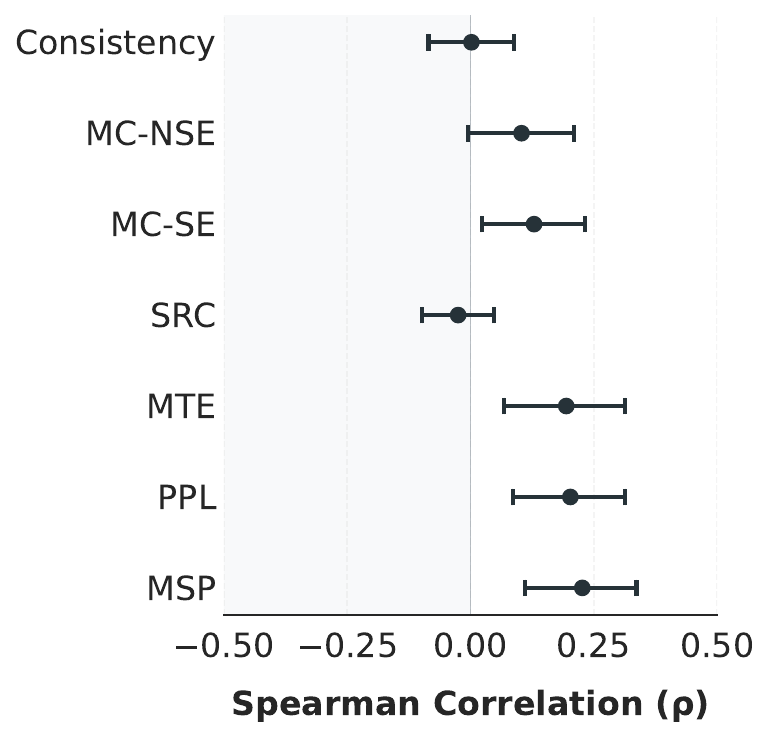}
        \caption{Answer Frequency \& Confidence Correlation \\ (incorrect samples)}
        \label{fig:answer_freq_corr_incorrect}
    \end{subfigure}%
    
    \caption{
    \textbf{Evidence for Implicit Answer Conditioning.}
    (a) Aggregated spearman correlations between reasoning length and confidence under CoT (Fisher z-transformed; 95\% CI). 
    Longer reasoning traces correspond to \emph{lower} confidence.
    (b) Confidence as a function of how often the final answer appears in the reasoning trace. Repeated answer mentions substantially increase answer-token likelihood.
    (c–d) Across datasets, answer frequency positively correlates with ATL-based uncertainty estimates. This correlation persists even for incorrect predictions, indicating correctness-agnostic confidence inflation. In contrast, Consistency and SRC show little sensitivity to answer frequency.
    }
    \label{fig:implicit_conditioning_merged}
\end{figure}

Figure~\ref{fig:implicit_conditioning_merged} summarizes evidence supporting implicit answer conditioning.
Panel (a) shows that reasoning length correlates negatively with confidence.
We then illustrate the conditioning effect directly on OK-VQA using Qwen3-VL-32B-IT with CoT prompting (\cref{fig:answer_freq_32b}). The open-ended format of OK-VQA allows unambiguous detection of answer mentions within the reasoning trace. Higher answer frequency consistently corresponds to higher confidence, even when considering only incorrect predictions (additional results are given in \suppref{J}).
Finally, to quantify the implicit answer conditioning effect, we compute Spearman correlations between answer frequency and confidence.
Across all datasets and models using CoT prompting (\cref{fig:answer_freq_corr_all,fig:answer_freq_corr_incorrect}), ATL-based UQ estimates show a consistent positive correlation that persists even when restricted to incorrect samples, confirming correctness-agnostic confidence inflation. In contrast, Consistency and SRC exhibit near-zero correlation.
Finally, an extreme form of implicit answer conditioning can be observed when models repeatedly reason and answer within the same conversational context, as we show in \suppref{K}.

\subsubsection{Masking Answers when Reasoning Improves UQ.}

As a proxy for implicit answer conditioning, we test whether explicit answer mentions drive confidence inflation. We perform a controlled intervention on generated CoT traces. If answer mentions in the reasoning trace increase the probability of the predicted answer, then removing them should reduce the confidence of the model.

For each sample, we replace every exact occurrence of the predicted answer in the reasoning trace with \texttt{[MASK]}, while keeping the final answer tokens unchanged. This intervention removes explicit answer mentions that serve as a proxy for conditioning strength. We then recompute the answer-token likelihoods by re-scoring the predicted answer sequence conditioned on the masked reasoning trace. Because the predicted answer is held fixed, accuracy is identical across the CoT and masked-CoT conditions by construction.
As a control, we also evaluate a \textit{Random Mask} baseline that masks an equivalent number of randomly selected reasoning tokens. This ensures that any improvements are not simply due to removing tokens in the reasoning trace.

OK-VQA’s open-ended format allows unambiguous detection of answer mentions. Since masking only affects explicit answer mentions, we restrict the analysis to the subset where the predicted answer appears in the reasoning trace ($n=3{,}900$).
As shown in \cref{tab:masking_cot_okvqa}, masking answer mentions from the reasoning trace substantially restores ranking quality for ATL-based estimates. For example, MSP PRR more than triples (0.141 → 0.475), with similar improvements for Perplexity and MTE, while Spearman correlation increases by 0.17–0.28. These improvements occur without any change in accuracy, indicating that answer presence in the reasoning context contributes directly to confidence inflation.
By contrast, random masking produces only moderate changes.
Masking does not fully recover the no-CoT baseline (0.475 vs.\ 0.498 PRR for MSP), suggesting that implicit answer conditioning persists beyond explicit mentions of the answer. 
We observe similar trends on MathVista (shown in \suppref{L}). However, in this setting random masking leads to a substantial drop in UQ ranking quality, suggesting that masking may disrupt tokens contributing to the structured reasoning required for mathematical solutions.

\begin{table}[tb]
\centering
\caption{\textbf{Masking answer mentions in CoT improves uncertainty.} Masked-CoT replaces occurrences of the predicted answer in the reasoning trace with \texttt{[MASK]}, while Random Masked-CoT masks an equal number of random reasoning tokens. Masking answer mentions substantially restores the ranking quality of ATL-based uncertainty estimates. Best values between CoT, Masked-CoT, and Random Masked-CoT are bold. Qwen3-VL-8B-IT on OK-VQA shown here, results for MathVista are given in \suppref{L}.}
\label{tab:masking_cot_okvqa}
\scriptsize
\setlength{\tabcolsep}{4pt}
\renewcommand{\arraystretch}{1.05}
\resizebox{\linewidth}{!}{
\begin{tabular}{l ccc ccc ccc ccc}
\toprule
 & \multicolumn{3}{c}{No CoT (Acc 59.7)} 
 & \multicolumn{3}{c}{CoT (Acc 57.7)}
 & \multicolumn{3}{c}{Masked-CoT (Acc 57.7)}
 & \multicolumn{3}{c}{Random Masked-CoT (Acc 57.7)} \\
\cmidrule(lr){2-4} 
\cmidrule(lr){5-7} 
\cmidrule(lr){8-10}
\cmidrule(lr){11-13}
UQ Method 
& AUGRC$\downarrow$ & Spear.$\uparrow$ & PRR$\uparrow$
& AUGRC$\downarrow$ & Spear.$\uparrow$ & PRR$\uparrow$
& AUGRC$\downarrow$ & Spear.$\uparrow$ & PRR$\uparrow$
& AUGRC$\downarrow$ & Spear.$\uparrow$ & PRR$\uparrow$ \\
\midrule
MSP & 0.152 & 0.375 & 0.498 
    & 0.196 & 0.115 & 0.141 
    & \textbf{0.161} & \textbf{0.378} & \textbf{0.475}
    & 0.190 & 0.161 & 0.210 \\

MTE & 0.154 & 0.364 & 0.481 
    & 0.204 & 0.057 & 0.071 
    & \textbf{0.181} & \textbf{0.230} & \textbf{0.329}
    & 0.198 & 0.102 & 0.137 \\

PPL & 0.156 & 0.345 & 0.469 
    & 0.204 & 0.059 & 0.075 
    & \textbf{0.166} & \textbf{0.340} & \textbf{0.436}
    & 0.197 & 0.113 & 0.146 \\
\bottomrule
\end{tabular}
}
\end{table}

\subsubsection{Computational Cost of Consistency.}
The robustness of consistency comes at a computational cost: unlike the
single-generation ATL estimates (MSP, PPL, MTE), which are read off a single
forward pass, consistency requires $K$ stochastic generations.
This overhead is shared by the Monte Carlo estimates (MC-SE, MC-NSE), but
unlike those, consistency remains reliable under reasoning, so the additional
samples translate into usable uncertainty signal rather than wasted compute.
The cost is most pronounced precisely where consistency is most useful: under
reasoning.
Two factors mitigate this in practice. The $K$ generations are independent and
can be batched, so wall-clock latency need not scale linearly with $K$ given
sufficient throughput, even though total token cost does; and $K$ can be tuned
to trade compute for ranking quality, with consistency degrading gracefully at
smaller $K$ (in \suppref{D}). Where multi-sample reasoning is
infeasible, SRC offers a cheaper
alternative that, while less consistently reliable, avoids the systematic
inflation that affects ATL estimates.

\FloatBarrier
\section{Conclusion}
\label{sec:conclusion}

We show that reasoning introduces a previously overlooked cost for uncertainty quantification in vision-language models. While reasoning improves task accuracy, it consistently degrades the ranking quality of many popular uncertainty estimates. Our analysis attributes this effect to implicit answer conditioning: reasoning traces semantically converge on a conclusion before the answer is generated, artificially inflating token-level confidence regardless of correctness. This phenomenon appears consistently across datasets, model families, and both prompted and reasoning-native models. Finally, we find that agreement-based consistency remains robust under reasoning and often improves, making it a practical uncertainty estimate for reasoning-enabled VLMs. 

\section*{Acknowledgements}
We acknowledge the computational resources provided by the National Academic Infrastructure for Supercomputing in Sweden (NAISS), funded by the Swedish Research Council, as well as the computational resources provided on the Berzelius system, funded by the Knut and Alice Wallenberg Foundation and operated by NAISS.
%
%
\bibliographystyle{splncs04}
\newpage
\bibliography{references}

@String(CVPR  = {IEEE Conf. Comput. Vis. Pattern Recog.})

@String(NeurIPS = {Adv. Neural Inform. Process. Syst.})

@String(ICLR  = {Int. Conf. Learn. Represent.})

@String(AAAI  = {AAAI})

@String(CVPR  = {CVPR})

@String(NeurIPS = {NeurIPS})

@String(ICLR  = {ICLR})

@article{bai2025qwen3,
  title={Qwen3-vl technical report},
  author={Bai, Shuai and Cai, Yuxuan and Chen, Ruizhe and Chen, Keqin and Chen, Xionghui and Cheng, Zesen and Deng, Lianghao and Ding, Wei and Gao, Chang and Ge, Chunjiang and others},
  journal={arXiv preprint arXiv:2511.21631},
  year={2025}
}

@article{ye2024benchmarking,
  title={Benchmarking llms via uncertainty quantification},
  author={Ye, Fanghua and Yang, Mingming and Pang, Jianhui and Wang, Longyue and Wong, Derek and Yilmaz, Emine and Shi, Shuming and Tu, Zhaopeng},
  journal={Advances in Neural Information Processing Systems},
  volume={37},
  pages={15356--15385},
  year={2024}
}

@inproceedings{srinivasan2024selective,
    title = "Selective ``Selective Prediction'': Reducing Unnecessary Abstention in Vision-Language Reasoning",
    author = "Srinivasan, Tejas  and
      Hessel, Jack  and
      Gupta, Tanmay  and
      Lin, Bill Yuchen  and
      Choi, Yejin  and
      Thomason, Jesse  and
      Chandu, Khyathi",
    editor = "Ku, Lun-Wei  and
      Martins, Andre  and
      Srikumar, Vivek",
    booktitle = "Findings of the Association for Computational Linguistics: ACL 2024",
    month = aug,
    year = "2024",
    address = "Bangkok, Thailand",
    publisher = "Association for Computational Linguistics",
    pages = "12935--12948",
}

@article{kostumov2024uncertainty,
  title={Uncertainty-aware evaluation for vision-language models},
  author={Kostumov, Vasily and Nutfullin, Bulat and Pilipenko, Oleg and Ilyushin, Eugene},
  journal={arXiv preprint arXiv:2402.14418},
  year={2024}
}

@article{kadavath2022language,
  title={Language models (mostly) know what they know},
  author={Kadavath, Saurav and Conerly, Tom and Askell, Amanda and Henighan, Tom and Drain, Dawn and Perez, Ethan and Schiefer, Nicholas and Hatfield-Dodds, Zac and DasSarma, Nova and Tran-Johnson, Eli and others},
  journal={arXiv preprint arXiv:2207.05221},
  year={2022}
}

@inproceedings{
    jaeger2022call,
    title={A Call to Reflect on Evaluation Practices for Failure Detection in Image Classification},
    author={Paul F Jaeger and Carsten Tim L{\"u}th and Lukas Klein and Till J. Bungert},
    booktitle={International Conference on Learning Representations},
    year={2023},
}

@inproceedings{wang2022self,
  author       = {Xuezhi Wang and
                  Jason Wei and
                  Dale Schuurmans and
                  Quoc V. Le and
                  Ed H. Chi and
                  Sharan Narang and
                  Aakanksha Chowdhery and
                  Denny Zhou},
  title        = {Self-Consistency Improves Chain of Thought Reasoning in Language Models},
  booktitle    = {The Eleventh International Conference on Learning Representations,
                  {ICLR} 2023, Kigali, Rwanda, May 1-5, 2023},
  year         = {2023},
  bibsource    = {dblp computer science bibliography, https://dblp.org}
}

@inproceedings{lyu2025calibrating,
  title={Calibrating large language models with sample consistency},
  author={Lyu, Qing and Shridhar, Kumar and Malaviya, Chaitanya and Zhang, Li and Elazar, Yanai and Tandon, Niket and Apidianaki, Marianna and Sachan, Mrinmaya and Callison-Burch, Chris},
  booktitle={Proceedings of the AAAI Conference on Artificial Intelligence},
  volume={39},
  pages={19260--19268},
  year={2025}
}

@article{wei2022chain,
  title={Chain-of-thought prompting elicits reasoning in large language models},
  author={Wei, Jason and Wang, Xuezhi and Schuurmans, Dale and Bosma, Maarten and Xia, Fei and Chi, Ed and Le, Quoc V and Zhou, Denny and others},
  journal={Advances in neural information processing systems},
  volume={35},
  pages={24824--24837},
  year={2022}
}

@article{traub2024overcoming,
  title={Overcoming common flaws in the evaluation of selective classification systems},
  author={Traub, Jeremias and Bungert, Till J and L{\"u}th, Carsten T and Baumgartner, Michael and Maier-Hein, Klaus H and Maier-Hein, Lena and J{\"a}ger, Paul F},
  journal={Advances in Neural Information Processing Systems},
  volume={37},
  pages={2323--2347},
  year={2024}
}

@InProceedings{ren2023self,
  title = 	 {Self-Evaluation Improves Selective Generation in Large Language Models},
  author =       {Ren, Jie and Zhao, Yao and Vu, Tu and Liu, Peter J. and Lakshminarayanan, Balaji},
  booktitle = 	 {Proceedings on "I Can't Believe It's Not Better: Failure  Modes in the Age of Foundation Models" at NeurIPS 2023 Workshops},
  pages = 	 {49--64},
  year = 	 {2023},
  editor = 	 {Antorán, Javier and Blaas, Arno and Buchanan, Kelly and Feng, Fan and Fortuin, Vincent and Ghalebikesabi, Sahra and Kriegler, Andreas and Mason, Ian and Rohde, David and Ruiz, Francisco J. R. and Uelwer, Tobias and Xie, Yubin and Yang, Rui},
  volume = 	 {239},
  series = 	 {Proceedings of Machine Learning Research},
  month = 	 {16 Dec},
  publisher =    {PMLR}
}

@inproceedings{geifman2018bias,
  author       = {Yonatan Geifman and
                  Guy Uziel and
                  Ran El{-}Yaniv},
  title        = {Bias-Reduced Uncertainty Estimation for Deep Neural Classifiers},
  booktitle    = {7th International Conference on Learning Representations, {ICLR} 2019,
                  New Orleans, LA, USA, May 6-9, 2019},
  year         = {2019},
  bibsource    = {dblp computer science bibliography, https://dblp.org}
}

@inproceedings{fadeeva2023lm,
  title={LM-polygraph: Uncertainty estimation for language models},
  author={Fadeeva, Ekaterina and Vashurin, Roman and Tsvigun, Akim and Vazhentsev, Artem and Petrakov, Sergey and Fedyanin, Kirill and Vasilev, Daniil and Goncharova, Elizaveta and Panchenko, Alexander and Panov, Maxim and others},
  booktitle={Proceedings of the 2023 Conference on Empirical Methods in Natural Language Processing: System Demonstrations},
  pages={446--461},
  year={2023}
}

@inproceedings{malinin2017incorporating,
  title={Incorporating uncertainty into deep learning for spoken language assessment},
  author={Malinin, Andrey and Ragni, Anton and Knill, Kate and Gales, Mark},
  booktitle={Proceedings of the 55th Annual Meeting of the Association for Computational Linguistics (Volume 2: Short Papers)},
  pages={45--50},
  year={2017}
}

@article{vashurin2025benchmarking,
  title={Benchmarking uncertainty quantification methods for large language models with lm-polygraph},
  author={Vashurin, Roman and Fadeeva, Ekaterina and Vazhentsev, Artem and Rvanova, Lyudmila and Vasilev, Daniil and Tsvigun, Akim and Petrakov, Sergey and Xing, Rui and Sadallah, Abdelrahman and Grishchenkov, Kirill and others},
  journal={Transactions of the Association for Computational Linguistics},
  volume={13},
  pages={220--248},
  year={2025}
}

@inproceedings{
yoon2025reasoning,
title={Reasoning Models Better Express Their Confidence},
author={Dongkeun Yoon and Seungone Kim and Sohee Yang and Sunkyoung Kim and Soyeon Kim and Yongil Kim and Eunbi Choi and Yireun Kim and Minjoon Seo},
booktitle={The Thirty-ninth Annual Conference on Neural Information Processing Systems},
year={2025},
}

@inproceedings{khan2024consistency,
  title={Consistency and uncertainty: Identifying unreliable responses from black-box vision-language models for selective visual question answering},
  author={Khan, Zaid and Fu, Yun},
  booktitle={Proceedings of the ieee/cvf conference on computer vision and pattern recognition},
  pages={10854--10863},
  year={2024}
}

@article{devic2025trace,
  title={Trace length is a simple uncertainty signal in reasoning models},
  author={Devic, Siddartha and Peale, Charlotte and Bradley, Arwen and Williamson, Sinead and Nakkiran, Preetum and Gollakota, Aravind},
  journal={arXiv preprint arXiv:2510.10409},
  year={2025}
}

@inproceedings{malinin2020uncertainty,
  title={Uncertainty Estimation in Autoregressive Structured Prediction},
  author={Malinin, Andrey and Gales, Mark},
  booktitle={International Conference on Learning Representations},
  year={2021}
}

@inproceedings{parkhi2012cats,
  title={Cats and dogs},
  author={Parkhi, Omkar M and Vedaldi, Andrea and Zisserman, Andrew and Jawahar, CV},
  booktitle={2012 IEEE conference on computer vision and pattern recognition},
  pages={3498--3505},
  year={2012},
  organization={IEEE}
}

@InProceedings{okvqa,
author = {Kenneth Marino and Mohammad Rastegari and Ali Farhadi and Roozbeh Mottaghi},
title = {OK-VQA: A Visual Question Answering Benchmark Requiring External Knowledge},
booktitle = {Conference on Computer Vision and Pattern Recognition (CVPR)},
year = {2019},
}

@inproceedings{yue2025mmmu,
  title={Mmmu-pro: A more robust multi-discipline multimodal understanding benchmark},
  author={Yue, Xiang and Zheng, Tianyu and Ni, Yuansheng and Wang, Yubo and Zhang, Kai and Tong, Shengbang and Sun, Yuxuan and Yu, Botao and Zhang, Ge and Sun, Huan and others},
  booktitle={Proceedings of the 63rd Annual Meeting of the Association for Computational Linguistics (Volume 1: Long Papers)},
  pages={15134--15186},
  year={2025}
}

@inproceedings{lu2023mathvista,
  author    = {Lu, Pan and Bansal, Hritik and Xia, Tony and Liu, Jiacheng and Li, Chunyuan and Hajishirzi, Hannaneh and Cheng, Hao and Chang, Kai-Wei and Galley, Michel and Gao, Jianfeng},
  title     = {MathVista: Evaluating Mathematical Reasoning of Foundation Models in Visual Contexts},
  booktitle={International Conference on Learning Representations (ICLR)},
  year      = {2024}
}

@article{lanham2023measuring,
  title={Measuring faithfulness in chain-of-thought reasoning},
  author={Lanham, Tamera and Chen, Anna and Radhakrishnan, Ansh and Steiner, Benoit and Denison, Carson and Hernandez, Danny and Li, Dustin and Durmus, Esin and Hubinger, Evan and Kernion, Jackson and others},
  journal={arXiv preprint arXiv:2307.13702},
  year={2023}
}

@ARTICLE{gemmateam2025gemma3technicalreport,
       author = {{Gemma Team} and {Kamath}, Aishwarya and {Ferret}, Johan and {Pathak}, Shreya and others},
        title = "{Gemma 3 Technical Report}",
      journal = {arXiv e-prints},
         year = 2025,
        month = mar,
          eid = {arXiv:2503.19786},
        pages = {arXiv:2503.19786},
archivePrefix = {arXiv},
       eprint = {2503.19786},
 primaryClass = {cs.CL},
       adsurl = {https://ui.adsabs.harvard.edu/abs/2025arXiv250319786G}
}

@article{kurz2025benchmarking,
  title={Benchmarking vision-language models for diagnostics in emergency and critical care settings},
  author={Kurz, Christoph F and Merzhevich, Tatiana and Eskofier, Bjoern M and Kather, Jakob Nikolas and Gmeiner, Benjamin},
  journal={npj Digital Medicine},
  volume={8},
  number={1},
  pages={423},
  year={2025},
  publisher={Nature Publishing Group UK London}
}

@inproceedings{xu2024vlm,
  title={Vlm-ad: End-to-end autonomous driving through vision-language model supervision},
  author={Xu, Yi and Hu, Yuxin and Zhang, Zaiwei and Meyer, Gregory P and Mustikovela, Siva Karthik and Srinivasa, Siddhartha and Wolff, Eric M and Huang, Xin},
  booktitle={Proceedings of the 9th Conference on Robot Learning (CoRL)},
  year={2025},
  address={Seoul, Korea},
}

@inproceedings{cole2023selectively,
  title={Selectively answering ambiguous questions},
  author={Cole, Jeremy and Zhang, Michael and Gillick, Dan and Eisenschlos, Julian and Dhingra, Bhuwan and Eisenstein, Jacob},
  booktitle={Proceedings of the 2023 Conference on Empirical Methods in Natural Language Processing},
  pages={530--543},
  year={2023}
}

@inproceedings{kuhn2023semantic,
  author       = {Lorenz Kuhn and
                  Yarin Gal and
                  Sebastian Farquhar},
  title        = {Semantic Uncertainty: Linguistic Invariances for Uncertainty Estimation
                  in Natural Language Generation},
  booktitle    = {The Eleventh International Conference on Learning Representations,
                  {ICLR} 2023, Kigali, Rwanda, May 1-5, 2023},
  year         = {2023},
  bibsource    = {dblp computer science bibliography, https://dblp.org}
}

@article{guo2025deepseek,
  title={DeepSeek-R1 incentivizes reasoning in LLMs through reinforcement learning},
  author={DeepSeek-AI},
  journal={Nature},
  year={2025},
  volume={645},
  pages={633 - 638},
}

@article{jaech2024openai,
  title={Openai o1 system card},
  author={Jaech, Aaron and Kalai, Adam and Lerer, Adam and Richardson, Adam and El-Kishky, Ahmed and Low, Aiden and Helyar, Alec and Madry, Aleksander and Beutel, Alex and Carney, Alex and others},
  journal={arXiv preprint arXiv:2412.16720},
  year={2024}
}

@article{kojima2022large,
  title={Large language models are zero-shot reasoners},
  author={Kojima, Takeshi and Gu, Shixiang Shane and Reid, Machel and Matsuo, Yutaka and Iwasawa, Yusuke},
  journal={Advances in neural information processing systems},
  volume={35},
  pages={22199--22213},
  year={2022}
}

@article{Zhu2025InternVL3EA,
  title={Internvl3: Exploring advanced training and test-time recipes for open-source multimodal models},
  author={Zhu, Jinguo and Wang, Weiyun and Chen, Zhe and Liu, Zhaoyang and Ye, Shenglong and Gu, Lixin and Tian, Hao and Duan, Yuchen and Su, Weijie and Shao, Jie and others},
  journal={arXiv preprint arXiv:2504.10479},
  year={2025}
}

\clearpage
\appendix
\section*{Appendix}
\appendix
\label{appendix}

\section{Expected Calibration Error}
\label{appendix:ece}
We additionally report the Expected Calibration Error (ECE) for every model and
dataset in \cref{tab:ece}. We restrict this analysis to the three 
measures bounded to $[0,1]$ (MSP, SRC, and sampled consistency) for which ECE is
well defined. Two trends hold across models. First, reasoning degrades the
calibration of MSP in the large majority of settings, sometimes sharply (e.g.,
InternVL3-9B on MathVista worsens from $0.047$ to $0.200$). Second, across all
sixteen model--dataset pairs for which a non-reasoning baseline is available,
reasoning improves the calibration of consistency. As a result,
the best-calibrated measure shifts from MSP without reasoning (lowest ECE in 13
of 16 settings) to consistency with reasoning (lowest in 13 of 16).

\begin{table}
\centering
\caption{Expected Calibration Error (ECE, 10 bins) across models and datasets. Lower is better; best per setting in \textbf{bold}.}
\label{tab:ece}
\setlength{\tabcolsep}{5pt}
\renewcommand{\arraystretch}{1.2}
\resizebox{\linewidth}{!}{%
\begin{tabular}{@{}ll rrr @{\hskip 16pt} rrr@{}}
\toprule
& & \multicolumn{3}{c}{\textbf{Non-Reasoning}} & \multicolumn{3}{c}{\textbf{Reasoning}} \\
\cmidrule(lr){3-5} \cmidrule(lr){6-8}
\textbf{Model} & \textbf{Dataset} & MSP & SRC & Cons. & MSP & SRC & Cons. \\
\midrule
\multirow{4}{*}{Gemma3-4B-IT} & Oxford-IIIT Pet & \bfseries 0.301 & 0.383 & 0.313 & \bfseries 0.161 & 0.287 & 0.161 \\
 & OK-VQA & \bfseries 0.349 & 0.416 & 0.450 & 0.410 & 0.557 & \bfseries 0.352 \\
 & MMMU-Pro-Vision & 0.581 & \bfseries 0.497 & 0.752 & 0.682 & \bfseries 0.366 & 0.542 \\
 & MathVista & \bfseries 0.397 & 0.585 & 0.521 & 0.513 & 0.440 & \bfseries 0.268 \\
\midrule
\multirow{4}{*}{Qwen3-VL-8B-IT} & Oxford-IIIT Pet & \bfseries 0.061 & 0.198 & 0.143 & 0.170 & 0.260 & \bfseries 0.103 \\
 & OK-VQA & \bfseries 0.199 & 0.330 & 0.227 & 0.270 & 0.382 & \bfseries 0.217 \\
 & MMMU-Pro-Vision & \bfseries 0.485 & 0.528 & 0.528 & 0.525 & 0.384 & \bfseries 0.336 \\
 & MathVista & \bfseries 0.214 & 0.326 & 0.242 & 0.224 & 0.233 & \bfseries 0.135 \\
\midrule
\multirow{4}{*}{Qwen3-VL-8B-Thinking} & Oxford-IIIT Pet & -- & -- & -- & 0.228 & 0.208 & \bfseries 0.117 \\
 & OK-VQA & -- & -- & -- & 0.341 & 0.339 & \bfseries 0.180 \\
 & MMMU-Pro-Vision & -- & -- & -- & 0.433 & 0.425 & \bfseries 0.284 \\
 & MathVista & -- & -- & -- & 0.189 & 0.194 & \bfseries 0.082 \\
\midrule
\multirow{4}{*}{Qwen3-VL-32B-IT} & Oxford-IIIT Pet & 0.150 & \bfseries 0.148 & 0.168 & 0.136 & 0.138 & \bfseries 0.116 \\
 & OK-VQA & \bfseries 0.213 & 0.316 & 0.310 & 0.358 & 0.391 & \bfseries 0.304 \\
 & MMMU-Pro-Vision & 0.431 & \bfseries 0.408 & 0.512 & 0.410 & 0.302 & \bfseries 0.283 \\
 & MathVista & \bfseries 0.174 & 0.218 & 0.234 & 0.182 & 0.130 & \bfseries 0.126 \\
\midrule
\multirow{4}{*}{InternVL3-9B} & Oxford-IIIT Pet & \bfseries 0.090 & 0.182 & 0.189 & 0.241 & 0.342 & \bfseries 0.150 \\
 & OK-VQA & \bfseries 0.061 & 0.284 & 0.223 & \bfseries 0.194 & 0.402 & 0.202 \\
 & MMMU-Pro-Vision & \bfseries 0.168 & 0.388 & 0.408 & 0.468 & 0.372 & \bfseries 0.354 \\
 & MathVista & \bfseries 0.047 & 0.247 & 0.179 & 0.200 & 0.243 & \bfseries 0.125 \\
\bottomrule
\end{tabular}
}
\end{table}

\FloatBarrier

\newpage
\section{Prompt Examples}
\label{appendix:prompt_examples}
\newcommand{\PromptSep}[1]{%
  \par\smallskip
  \noindent\textbf{#1}\par
  \noindent\rule{\linewidth}{0.4pt}%
  \par\smallskip
}

\newtcolorbox{PromptBox}[1]{%
  breakable,
  boxrule=0.5pt,
  colback=gray!5,
  colframe=black!70,
  arc=2pt,
  left=6pt,
  right=6pt,
  top=4pt,
  bottom=4pt,
  title={#1},
  fonttitle=\bfseries
}

\lstset{
  basicstyle=\ttfamily\small,
  breaklines=true,
  columns=fullflexible,
  showstringspaces=false,
  xleftmargin=0pt
}


\needspace{8\baselineskip}
\begin{PromptBox}{Standard VQA Prompt (Without vs. With CoT)}

\PromptSep{Shared context (both)}
\begin{lstlisting}
You are a VQA assistant. You must answer the given question based on 
the image and the provided multiple-choice options. Only one of the 
options is correct.

<image>
Question: {question}
Options: {options}
\end{lstlisting}

\PromptSep{Without CoT answer mode only}
\begin{lstlisting}
Respond ONLY in the following wrapped format:
<answer><option></answer>
\end{lstlisting}

\PromptSep{With CoT only}
\begin{lstlisting}
First, think step by step and briefly describe the cues you would use 
to select the correct option, then answer.

Respond ONLY in the following wrapped format:
<thought> your thought process </thought>
<answer><option></answer>
\end{lstlisting}

\PromptSep{Format rules (both)}
\begin{lstlisting}
Replace <option> with the chosen option text, copied verbatim from 
the list of options. Include both opening and closing tags exactly 
once. IMPORTANT: Any deviation from this exact tag format is 
incorrect.
\end{lstlisting}

\end{PromptBox}


\needspace{8\baselineskip}
\begin{PromptBox}{Self-Reported Confidence Prompt (Without vs. With CoT)}

\PromptSep{Shared context (both)}
\begin{lstlisting}
You are a VQA assistant.

<image>
You previously answered the question: {question} with the answer:
{answer} out of the possible options: {mc_options}.

Your task is now to self-report your certainty about your answer.

\end{lstlisting}

\PromptSep{Without CoT answer mode only}
\begin{lstlisting}
Are you confident in your answer?
Respond ONLY in the following wrapped format:

<answer>yes</answer> or <answer>no</answer> 
\end{lstlisting}

\PromptSep{With CoT only}
\begin{lstlisting}
First, think step by step and briefly explain why you are or are not 
confident in your answer. Then, answer: Are you confident in your 
answer?

Respond ONLY in the following wrapped format:

<thought> your reasoning </thought>
<answer>yes</answer> or <answer>no</answer> 

\end{lstlisting}

\PromptSep{Format rules (both)}
\begin{lstlisting}
Do not add any text outside the tags.
Do not add spaces or punctuation inside the yes/no tags.
\end{lstlisting}

\end{PromptBox}

\needspace{10\baselineskip}
\begin{PromptBox}{Dataset Instructions}

\PromptSep{OK-VQA}
\begin{lstlisting}
When giving the final answer, give a short, direct answer using 1-2 
words. Use the most common, canonical name.

Examples:

    Q: What sort of vehicles uses this item?
    A: <answer>firetruck</answer>

    Q: How many chromosomes do these creatures have?
    A: <answer>23</answer>

    Q: What material is this vase made out of?
    A: <answer>ceramic</answer>

    Q: What is the man doing?
    A: <answer>standing</answer>
\end{lstlisting}

\PromptSep{MathVista}
\begin{lstlisting}
If options are listed in the question, choose one option and return 
exactly that option text in the final answer. 
Do not include equations, units, or explanation in the final answer.
\end{lstlisting}

\end{PromptBox}

\section{Decoding parameters}
\label{appendix:decoding_parameters}
For the all models, we uniformly apply the decoding parameters detailed in \cref{tab:decoding_params} across all uncertainty measures and experiments. Decoding parameters for Qwen3-VL models are set following the recommended 
settings reported in the Qwen3-VL technical report. For Gemma-3, the technical report does not prescribe decoding parameters; we therefore adopt the default recommended settings used in Google’s public developer interfaces and documentation. We set the maximum generated tokens sufficiently large (20k) to ensure that no generated responses are truncated for any experiments.

\begin{table}
\centering
\caption{Decoding hyperparameters used for each model family.}
\label{tab:decoding_params}
\resizebox{\linewidth}{!}{%
\begin{tabular}{lcccc}
\toprule
\textbf{Model} & \textbf{Temperature} & \textbf{Top-$p$} & \textbf{Top-$k$}  & \textbf{Presence Penalty} \\
\midrule
Gemma3-4B-IT & 1.0 & 0.95 & 64 & 1.0 \\
InternVL3-9B & 0.8 & 0.80 & 40 & 0.0 \\
Qwen3-VL-8B-IT & 1.0 & 1.0 & 40 & 2.0 \\
Qwen3-VL-8B-Thinking & 1.0 & 0.95 & 20 & 1.5 \\
Qwen3-VL-32B-IT & 0.7 & 0.80 & 20 & 1.5 \\
\bottomrule
\end{tabular}
}
\end{table}

\FloatBarrier

\section{K-Ablation}
\label{appendix:K_ablation}
We conduct an ablation study on MathVista across different values of k (the number of sampled answers or reasoning traces) using CoT prompting with Qwen3-VL-8B-Instruct. Increasing k consistently improves agreement-based consistency, which benefits from additional sampled reasoning paths. In contrast, the effect of increasing k on MC-SE and MC-NSE is less consistent: while modest improvements are observed at larger k, their performance varies across intermediate values. Overall, these results suggest that agreement-based consistency more reliably benefits from additional sampling than token-likelihood–based Monte Carlo estimates. We note that between runs, the predictive performance changes, as the final answer is determined by majority vote.

\begin{figure}[!ht]
    \centering
    
    \begin{subfigure}[t]{0.48\linewidth}
        \centering
        \includegraphics[width=\linewidth]{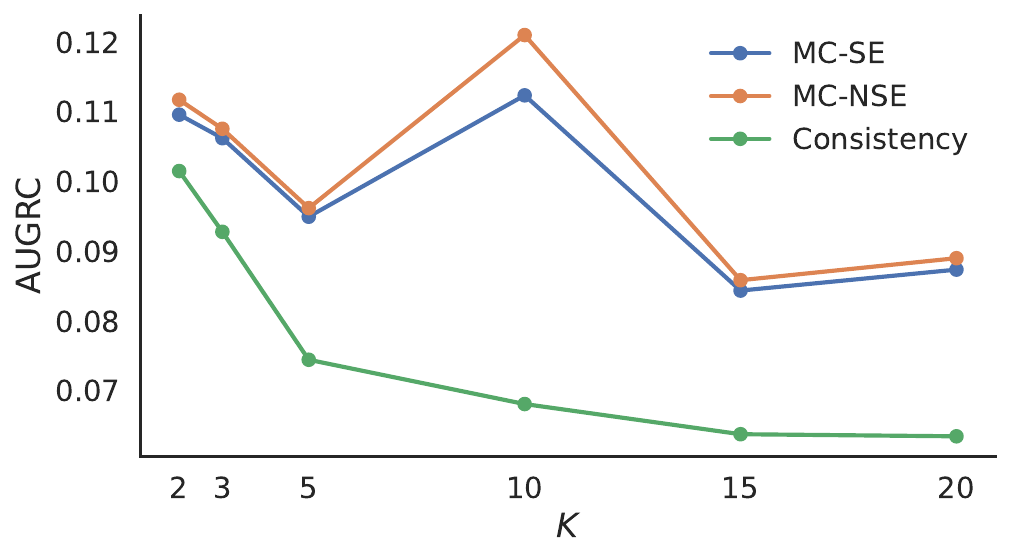}
        \caption{AUGRC across $k$ values.}
        \label{fig:augrc_vs_k}
    \end{subfigure}
    \hfill
    \begin{subfigure}[t]{0.48\linewidth}
        \centering
        \includegraphics[width=\linewidth]{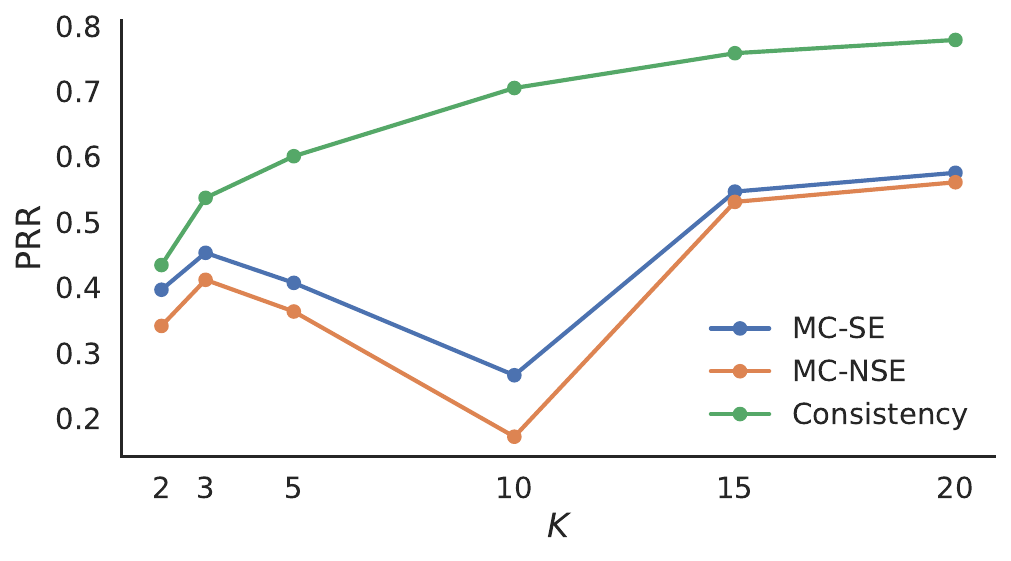}
        \caption{PRR across $k$ values.}
        \label{fig:prr_vs_k}
    \end{subfigure}
    
    \caption{Effect of sample size $k$ on uncertainty quality. Lower AUGRC is better, higher PRR is better.}
    \label{fig:k_scaling}
\end{figure}

\section{Answer Extraction}
\label{appendix:answer_extraction}
Models are prompted to wrap their final answer in \texttt{<answer>...</answer>} tags. We locate this span in token space by exact subsequence matching on token IDs, checking \texttt{<answer>} and \texttt{</answer>} against tokenizations of each marker preceded by no prefix, a space, a newline, a tab, \texttt{\textbackslash r}, or \texttt{\textbackslash r\textbackslash n}, since tokenizers can fuse leading whitespace into the marker. If \texttt{</answer>} is not found after the start marker, we fall back to the chat template's turn delimiter (\texttt{<end\_of\_turn>}, under the same whitespace-prefix variants); if that is also absent, we take the end of the sequence (after trimming trailing special tokens). As a last resort, if token-level matching fails entirely (e.g.\ the markers are fused with adjacent text into a single token), we fall back to a regex match on the decoded text and map the matched character span back to token indices.

\newpage
\section{Selective Generation Results on all datasets}
\label{appendix:all_selective_generation_tables}
\FloatBarrier

\begin{table}[!ht]
\centering
\small
\setlength{\tabcolsep}{6pt}
\renewcommand{\arraystretch}{1.2}
\caption{Selective Generation Results on Oxford Pets}
\label{tab:oxfordpets_all_models}
\resizebox{\linewidth}{!}{%
\begin{tabular}{ll ccc ccc ccc}
\toprule
 & & \multicolumn{3}{c}{\textbf{No CoT}} & \multicolumn{3}{c}{\textbf{CoT}} & \multicolumn{3}{c}{\textbf{$\Delta$ (CoT$-$ No CoT)}} \\
\cmidrule(lr){3-5} \cmidrule(lr){6-8} \cmidrule(lr){9-11}
\textbf{Type} & \textbf{UQ Estimate} & \small{AUGRC$\downarrow$} & \small{PRR$\uparrow$} & \small{Spear.$\uparrow$} & \small{AUGRC$\downarrow$} & \small{PRR$\uparrow$} & \small{Spear.$\uparrow$} & \small{AUGRC} & \small{PRR} & \small{Spear.} \\
\midrule
\rowcolor{gray!5} \multicolumn{8}{l}{\textbf{Gemma3-4B-IT}} \\
\multirow{4}{*}{\shortstack[l]{\textit{Single-Generation} \\ \scriptsize (Acc: 65.2 / 70.4)}} & MSP & 0.122 & 0.498 & 0.380 & 0.139 & 0.200 & 0.068 & 0.018 & -0.297 & -0.312 \\
 & PPL & 0.121 & 0.498 & 0.384 & 0.141 & 0.186 & 0.054 & 0.020 & -0.312 & -0.330 \\
 & MTE & \textbf{0.118} & \textbf{0.521} & \textbf{0.404} & 0.136 & 0.219 & 0.090 & 0.018 & -0.302 & -0.314 \\
 & SRC & 0.121 & 0.511 & 0.388 & \textbf{0.104} & \textbf{0.504} & \textbf{0.333} & -0.016 & -0.007 & -0.055 \\
\addlinespace[0.3em] \cmidrule{2-11} \addlinespace[0.3em]
\multirow{3}{*}{\shortstack[l]{\textit{Multi-Generation} \\ \scriptsize (Acc: 65.4 / 71.8)}} & MC-SE & 0.122 & 0.497 & 0.375 & 0.123 & \textbf{0.296} & 0.136 & 0.002 & -0.201 & -0.240 \\
 & MC-NSE & \textbf{0.121} & \textbf{0.498} & \textbf{0.380} & 0.125 & 0.288 & 0.123 & 0.004 & -0.210 & -0.257 \\
 & Consistency & 0.147 & 0.475 & 0.295 & \textbf{0.112} & 0.283 & \textbf{0.313} & -0.034 & -0.192 & 0.018 \\
 \midrule
\rowcolor{gray!5} \multicolumn{8}{l}{\textbf{InternVL3-9B}} \\
\multirow{4}{*}{\shortstack[l]{\textit{Single-Generation} \\ \scriptsize (Acc: 63.1 / 65.5)}} & MSP & \textbf{0.124} & \textbf{0.624} & \textbf{0.435} & 0.170 & 0.078 & 0.015 & 0.046 & -0.546 & -0.420 \\
 & PPL & 0.127 & 0.613 & 0.415 & 0.173 & 0.047 & -0.007 & 0.047 & -0.566 & -0.422 \\
 & MTE & 0.126 & 0.619 & 0.419 & 0.170 & 0.064 & 0.015 & 0.044 & -0.554 & -0.404 \\
 & SRC & 0.130 & 0.612 & 0.392 & \textbf{0.127} & \textbf{0.513} & \textbf{0.331} & -0.003 & -0.099 & -0.061 \\
\addlinespace[0.3em] \cmidrule{2-11} \addlinespace[0.3em]
\multirow{3}{*}{\shortstack[l]{\textit{Multi-Generation} \\ \scriptsize (Acc: 64.2 / 67.0)}} & MC-SE & \textbf{0.120} & \textbf{0.621} & \textbf{0.424} & 0.156 & 0.202 & 0.071 & 0.035 & -0.419 & -0.353 \\
 & MC-NSE & 0.124 & 0.605 & 0.400 & 0.158 & 0.183 & 0.056 & 0.034 & -0.422 & -0.345 \\
 & Consistency & 0.145 & 0.448 & 0.355 & \textbf{0.131} & \textbf{0.343} & \textbf{0.302} & -0.014 & -0.106 & -0.053 \\
\midrule \rowcolor{gray!5} \multicolumn{8}{l}{\textbf{Qwen3-VL-32B-IT}} \\
\multirow{4}{*}{\shortstack[l]{\textit{Single-Generation} \\ \scriptsize (Acc: 82.2 / 86.4)}} & MSP & 0.046 & 0.644 & 0.386 & 0.076 & 0.010 & -0.082 & 0.030 & -0.635 & -0.467 \\
 & PPL & 0.049 & 0.534 & 0.395 & 0.076 & 0.012 & -0.082 & 0.027 & -0.522 & -0.477 \\
 & MTE & 0.046 & 0.683 & 0.392 & 0.076 & 0.016 & -0.080 & 0.030 & -0.668 & -0.472 \\
 & SRC & \textbf{0.039} & \textbf{0.770} & \textbf{0.457} & \textbf{0.048} & \textbf{0.319} & \textbf{0.289} & 0.009 & -0.451 & -0.168 \\
\addlinespace[0.3em] \cmidrule{2-11} \addlinespace[0.3em]
\multirow{3}{*}{\shortstack[l]{\textit{Multi-Generation} \\ \scriptsize (Acc: 81.8 / 86.4)}} & MC-SE & \textbf{0.047} & \textbf{0.659} & \textbf{0.399} & 0.076 & -0.002 & -0.077 & 0.029 & -0.661 & -0.476 \\
 & MC-NSE & \textbf{0.047} & \textbf{0.659} & \textbf{0.399} & 0.076 & -0.001 & -0.077 & 0.029 & -0.660 & -0.476 \\
 & Consistency & 0.081 & 0.382 & 0.222 & \textbf{0.065} & \textbf{0.328} & \textbf{0.346} & -0.016 & -0.054 & 0.124 \\
\bottomrule
\end{tabular}
}
\end{table}

\begin{table}[!ht]
\centering
\small
\setlength{\tabcolsep}{6pt}
\renewcommand{\arraystretch}{1.2}
\caption{Selective Generation Results on OK-VQA}
\label{tab:OKVQA_all_models}
\resizebox{\linewidth}{!}{%
\begin{tabular}{ll ccc ccc ccc}
\toprule
 & & \multicolumn{3}{c}{\textbf{No CoT}} & \multicolumn{3}{c}{\textbf{CoT}} & \multicolumn{3}{c}{\textbf{$\Delta$ (CoT$-$ No CoT)}} \\
\cmidrule(lr){3-5} \cmidrule(lr){6-8} \cmidrule(lr){9-11}
\textbf{Type} & \textbf{UQ Estimate} & \small{AUGRC$\downarrow$} & \small{PRR$\uparrow$} & \small{Spear.$\uparrow$} & \small{AUGRC$\downarrow$} & \small{PRR$\uparrow$} & \small{Spear.$\uparrow$} & \small{AUGRC} & \small{PRR} & \small{Spear.} \\
\midrule
\rowcolor{gray!5} \multicolumn{8}{l}{\textbf{Gemma3-4B-IT}} \\
\multirow{4}{*}{\shortstack[l]{\textit{Single-Generation} \\ \scriptsize (Acc: 38.8 / 35.4)}} & MSP & 0.270 & \textbf{0.333} & 0.266 & 0.294 & 0.244 & 0.218 & 0.024 & -0.090 & -0.048 \\
 & PPL & 0.278 & 0.291 & 0.207 & 0.305 & 0.188 & 0.135 & 0.027 & -0.103 & -0.072 \\
 & MTE & 0.276 & 0.297 & 0.222 & 0.303 & 0.192 & 0.152 & 0.027 & -0.105 & -0.071 \\
 & SRC & \textbf{0.270} & 0.328 & \textbf{0.268} & \textbf{0.289} & \textbf{0.304} & \textbf{0.261} & 0.018 & -0.025 & -0.007 \\
\addlinespace[0.3em] \cmidrule{2-11} \addlinespace[0.3em]
\multirow{3}{*}{\shortstack[l]{\textit{Multi-Generation} \\ \scriptsize (Acc: 39.0 / 37.0)}} & MC-SE & \textbf{0.270} & \textbf{0.336} & 0.262 & 0.285 & 0.279 & 0.226 & 0.015 & -0.057 & -0.036 \\
 & MC-NSE & 0.279 & 0.290 & 0.198 & 0.298 & 0.213 & 0.130 & 0.019 & -0.077 & -0.068 \\
 & Consistency & 0.273 & 0.239 & \textbf{0.274} & \textbf{0.264} & \textbf{0.376} & \textbf{0.392} & -0.008 & 0.137 & 0.118 \\
 \midrule \rowcolor{gray!5} \multicolumn{8}{l}{\textbf{InternVL3-9B}} \\
\multirow{4}{*}{\shortstack[l]{\textit{Single-Generation} \\ \scriptsize (Acc: 54.3 / 43.0)}} & MSP & \textbf{0.154} & \textbf{0.682} & \textbf{0.557} & \textbf{0.220} & \textbf{0.539} & \textbf{0.479} & 0.066 & -0.144 & -0.078 \\
 & PPL & 0.165 & 0.615 & 0.474 & 0.235 & 0.442 & 0.367 & 0.070 & -0.173 & -0.107 \\
 & MTE & 0.169 & 0.594 & 0.447 & 0.238 & 0.425 & 0.345 & 0.070 & -0.168 & -0.102 \\
 & SRC & 0.189 & 0.385 & 0.298 & 0.245 & 0.354 & 0.294 & 0.057 & -0.031 & -0.004 \\
\addlinespace[0.3em] \cmidrule{2-11} \addlinespace[0.3em]
\multirow{3}{*}{\shortstack[l]{\textit{Multi-Generation} \\ \scriptsize (Acc: 55.4 / 47.2)}} & MC-SE & \textbf{0.152} & \textbf{0.666} & \textbf{0.532} & 0.201 & \textbf{0.566} & 0.462 & 0.049 & -0.100 & -0.070 \\
 & MC-NSE & 0.161 & 0.610 & 0.466 & 0.215 & 0.477 & 0.363 & 0.054 & -0.133 & -0.103 \\
 & Consistency & 0.167 & 0.460 & 0.451 & \textbf{0.199} & 0.545 & \textbf{0.486} & 0.032 & 0.086 & 0.036 \\
\midrule \rowcolor{gray!5} \multicolumn{8}{l}{\textbf{Qwen3-VL-32B-IT}} \\
\multirow{4}{*}{\shortstack[l]{\textit{Single-Generation} \\ \scriptsize (Acc: 57.9 / 51.8)}} & MSP & \textbf{0.156} & \textbf{0.536} & \textbf{0.417} & 0.196 & \textbf{0.394} & 0.335 & 0.040 & -0.141 & -0.083 \\
 & PPL & 0.158 & 0.519 & 0.402 & 0.199 & 0.359 & 0.309 & 0.041 & -0.160 & -0.093 \\
 & MTE & 0.156 & 0.526 & 0.413 & 0.199 & 0.358 & 0.308 & 0.043 & -0.168 & -0.105 \\
 & SRC & 0.168 & 0.358 & 0.341 & \textbf{0.195} & 0.391 & \textbf{0.352} & 0.028 & 0.033 & 0.011 \\
\addlinespace[0.3em] \cmidrule{2-11} \addlinespace[0.3em]
\multirow{3}{*}{\shortstack[l]{\textit{Multi-Generation} \\ \scriptsize (Acc: 58.1 / 52.9)}} & MC-SE & \textbf{0.156} & \textbf{0.530} & \textbf{0.407} & \textbf{0.186} & \textbf{0.463} & 0.368 & 0.030 & -0.067 & -0.039 \\
 & MC-NSE & 0.158 & 0.514 & 0.392 & 0.189 & 0.434 & 0.344 & 0.031 & -0.080 & -0.048 \\
 & Consistency & 0.182 & 0.225 & 0.264 & 0.190 & 0.346 & \textbf{0.372} & 0.007 & 0.121 & 0.108 \\
\bottomrule
\end{tabular}
}
\end{table}

\begin{table}[!ht]
\centering
\small
\setlength{\tabcolsep}{6pt}
\renewcommand{\arraystretch}{1.2}
\caption{Selective Generation Results on MMMU-Pro-Vision}
\label{tab:MMMU_vision_all_models}
\resizebox{\linewidth}{!}{%
\begin{tabular}{ll ccc ccc ccc}
\toprule
 & & \multicolumn{3}{c}{\textbf{No CoT}} & \multicolumn{3}{c}{\textbf{CoT}} & \multicolumn{3}{c}{\textbf{$\Delta$ (CoT$-$ No CoT)}} \\
\cmidrule(lr){3-5} \cmidrule(lr){6-8} \cmidrule(lr){9-11}
\textbf{Type} & \textbf{UQ Estimate} & \small{AUGRC$\downarrow$} & \small{PRR$\uparrow$} & \small{Spear.$\uparrow$} & \small{AUGRC$\downarrow$} & \small{PRR$\uparrow$} & \small{Spear.$\uparrow$} & \small{AUGRC} & \small{PRR} & \small{Spear.} \\
\midrule
\rowcolor{gray!5} \multicolumn{8}{l}{\textbf{Gemma3-4B-IT}} \\
\multirow{4}{*}{\shortstack[l]{\textit{Single-Generation} \\ \scriptsize (Acc: 19.0 / 18.9)}} & MSP & 0.401 & 0.118 & 0.037 & 0.398 & 0.036 & 0.069 & -0.003 & -0.082 & 0.032 \\
 & PPL & 0.401 & 0.116 & 0.035 & 0.398 & 0.036 & 0.069 & -0.003 & -0.081 & 0.034 \\
 & MTE & \textbf{0.398} & \textbf{0.129} & \textbf{0.059} & 0.397 & 0.039 & 0.075 & -0.001 & -0.090 & 0.016 \\
 & SRC & 0.400 & 0.077 & 0.047 & \textbf{0.394} & \textbf{0.122} & \textbf{0.108} & -0.006 & 0.045 & 0.060 \\
\addlinespace[0.3em] \cmidrule{2-11} \addlinespace[0.3em]
\multirow{3}{*}{\shortstack[l]{\textit{Multi-Generation} \\ \scriptsize (Acc: 18.6 / 20.3)}} & MC-SE & \textbf{0.401} & \textbf{0.116} & \textbf{0.056} & 0.390 & 0.072 & 0.076 & -0.011 & -0.044 & 0.019 \\
 & MC-NSE & 0.401 & 0.115 & 0.056 & 0.390 & 0.071 & 0.075 & -0.011 & -0.044 & 0.019 \\
 & Consistency & 0.406 & 0.016 & 0.042 & \textbf{0.385} & \textbf{0.088} & \textbf{0.123} & -0.021 & 0.072 & 0.081 \\
 \midrule \rowcolor{gray!5} \multicolumn{8}{l}{\textbf{InternVL3-9B}} \\
\multirow{4}{*}{\shortstack[l]{\textit{Single-Generation} \\ \scriptsize (Acc: 28.9 / 29.8)}} & MSP & 0.312 & 0.425 & 0.332 & 0.330 & 0.143 & 0.159 & 0.018 & -0.282 & -0.173 \\
 & PPL & 0.312 & 0.425 & 0.332 & 0.330 & 0.143 & 0.159 & 0.018 & -0.282 & -0.173 \\
 & MTE & \textbf{0.308} & \textbf{0.452} & \textbf{0.361} & \textbf{0.330} & 0.148 & \textbf{0.162} & 0.022 & -0.304 & -0.199 \\
 & SRC & 0.333 & 0.222 & 0.174 & 0.330 & \textbf{0.212} & 0.160 & -0.003 & -0.010 & -0.014 \\
\addlinespace[0.3em] \cmidrule{2-11} \addlinespace[0.3em]
\multirow{3}{*}{\shortstack[l]{\textit{Multi-Generation} \\ \scriptsize (Acc: 29.4 / 31.4)}} & MC-SE & \textbf{0.309} & \textbf{0.440} & \textbf{0.335} & 0.312 & 0.268 & 0.238 & 0.002 & -0.172 & -0.098 \\
 & MC-NSE & \textbf{0.309} & \textbf{0.440} & \textbf{0.335} & 0.312 & 0.268 & 0.238 & 0.002 & -0.172 & -0.098 \\
 & Consistency & 0.316 & 0.316 & 0.283 & \textbf{0.302} & \textbf{0.338} & \textbf{0.307} & -0.014 & 0.022 & 0.024 \\
\midrule \rowcolor{gray!5} \multicolumn{8}{l}{\textbf{Qwen3-VL-32B-IT}} \\
\multirow{4}{*}{\shortstack[l]{\textit{Single-Generation} \\ \scriptsize (Acc: 42.9 / 56.7)}} & MSP & 0.227 & 0.521 & 0.415 & 0.198 & 0.124 & 0.210 & -0.028 & -0.397 & -0.205 \\
 & PPL & 0.227 & 0.521 & 0.415 & 0.198 & 0.124 & 0.210 & -0.028 & -0.397 & -0.205 \\
 & MTE & \textbf{0.225} & \textbf{0.544} & \textbf{0.424} & \textbf{0.181} & 0.303 & \textbf{0.245} & -0.044 & -0.242 & -0.179 \\
 & SRC & 0.238 & 0.459 & 0.336 & 0.189 & \textbf{0.305} & 0.193 & -0.049 & -0.154 & -0.143 \\
\addlinespace[0.3em] \cmidrule{2-11} \addlinespace[0.3em]
\multirow{3}{*}{\shortstack[l]{\textit{Multi-Generation} \\ \scriptsize (Acc: 43.8 / 58.6)}} & MC-SE & \textbf{0.226} & \textbf{0.481} & \textbf{0.387} & 0.180 & 0.201 & 0.229 & -0.046 & -0.280 & -0.158 \\
 & MC-NSE & \textbf{0.226} & \textbf{0.481} & \textbf{0.387} & 0.179 & 0.201 & 0.229 & -0.046 & -0.279 & -0.158 \\
 & Consistency & 0.263 & 0.059 & 0.214 & \textbf{0.168} & \textbf{0.295} & \textbf{0.322} & -0.094 & 0.236 & 0.107 \\
\bottomrule
\end{tabular}
}
\end{table}

\begin{table}[!ht]
\centering
\small
\setlength{\tabcolsep}{6pt}
\renewcommand{\arraystretch}{1.2}
\caption{Selective Generation Results on MathVista}
\label{tab:MathVista_vision_all_models}
\resizebox{\linewidth}{!}{%
\begin{tabular}{ll ccc ccc ccc}
\toprule
 & & \multicolumn{3}{c}{\textbf{No CoT}} & \multicolumn{3}{c}{\textbf{CoT}} & \multicolumn{3}{c}{\textbf{$\Delta$ (CoT$-$ No CoT)}} \\
\cmidrule(lr){3-5} \cmidrule(lr){6-8} \cmidrule(lr){9-11}
\textbf{Type} & \textbf{UQ Estimate} & \small{AUGRC$\downarrow$} & \small{PRR$\uparrow$} & \small{Spear.$\uparrow$} & \small{AUGRC$\downarrow$} & \small{PRR$\uparrow$} & \small{Spear.$\uparrow$} & \small{AUGRC} & \small{PRR} & \small{Spear.} \\
\midrule
\rowcolor{gray!5} \multicolumn{8}{l}{\textbf{Gemma3-4B-IT}} \\
\multirow{4}{*}{\shortstack[l]{\textit{Single-Generation} \\ \scriptsize (Acc: 34.7 / 41.9)}} & MSP & \textbf{0.292} & \textbf{0.317} & \textbf{0.256} & \textbf{0.269} & \textbf{0.202} & \textbf{0.150} & -0.022 & -0.115 & -0.106 \\
 & PPL & 0.300 & 0.239 & 0.192 & 0.275 & 0.152 & 0.110 & -0.025 & -0.087 & -0.082 \\
 & MTE & 0.302 & 0.229 & 0.182 & 0.276 & 0.152 & 0.107 & -0.026 & -0.077 & -0.074 \\
 & SRC & 0.340 & 0.022 & -0.094 & 0.289 & 0.083 & 0.012 & -0.051 & 0.060 & 0.106 \\
\addlinespace[0.3em] \cmidrule{2-11} \addlinespace[0.3em]
\multirow{3}{*}{\shortstack[l]{\textit{Multi-Generation} \\ \scriptsize (Acc: 35.2 / 43.9)}} & MC-SE & \textbf{0.289} & \textbf{0.318} & \textbf{0.252} & 0.264 & 0.217 & 0.120 & -0.026 & -0.100 & -0.132 \\
 & MC-NSE & 0.300 & 0.238 & 0.178 & 0.268 & 0.197 & 0.092 & -0.032 & -0.041 & -0.086 \\
 & Consistency & 0.296 & 0.237 & 0.249 & \textbf{0.230} & \textbf{0.441} & \textbf{0.359} & -0.066 & 0.203 & 0.111 \\
 \midrule \rowcolor{gray!5} \multicolumn{8}{l}{\textbf{InternVL3-9B}} \\
\multirow{4}{*}{\shortstack[l]{\textit{Single-Generation} \\ \scriptsize (Acc: 63.7 / 62.6)}} & MSP & \textbf{0.103} & \textbf{0.747} & \textbf{0.567} & 0.162 & 0.153 & 0.179 & 0.059 & -0.593 & -0.389 \\
 & PPL & 0.113 & 0.689 & 0.495 & 0.171 & 0.108 & 0.117 & 0.058 & -0.581 & -0.378 \\
 & MTE & 0.111 & 0.689 & 0.508 & 0.167 & 0.124 & 0.143 & 0.056 & -0.565 & -0.364 \\
 & SRC & 0.141 & 0.487 & 0.295 & \textbf{0.145} & \textbf{0.450} & \textbf{0.304} & 0.004 & -0.038 & 0.010 \\
\addlinespace[0.3em] \cmidrule{2-11} \addlinespace[0.3em]
\multirow{3}{*}{\shortstack[l]{\textit{Multi-Generation} \\ \scriptsize (Acc: 66.0 / 66.0)}} & MC-SE & \textbf{0.096} & \textbf{0.727} & \textbf{0.544} & 0.135 & 0.401 & 0.260 & 0.039 & -0.327 & -0.283 \\
 & MC-NSE & 0.106 & 0.669 & 0.469 & 0.149 & 0.331 & 0.159 & 0.042 & -0.338 & -0.310 \\
 & Consistency & 0.110 & 0.532 & 0.484 & \textbf{0.101} & \textbf{0.699} & \textbf{0.520} & -0.009 & 0.167 & 0.036 \\
\midrule \rowcolor{gray!5} \multicolumn{8}{l}{\textbf{Qwen3-VL-32B-IT}} \\
\multirow{4}{*}{\shortstack[l]{\textit{Single-Generation} \\ \scriptsize (Acc: 70.9 / 80.8)}} & MSP & \textbf{0.095} & \textbf{0.627} & \textbf{0.388} & 0.071 & 0.338 & 0.218 & -0.023 & -0.289 & -0.170 \\
 & PPL & 0.100 & 0.586 & 0.347 & 0.075 & 0.289 & 0.190 & -0.025 & -0.297 & -0.158 \\
 & MTE & 0.100 & 0.585 & 0.349 & 0.075 & 0.292 & 0.184 & -0.024 & -0.293 & -0.165 \\
 & SRC & 0.096 & 0.491 & 0.377 & \textbf{0.048} & \textbf{0.711} & \textbf{0.425} & -0.048 & 0.220 & 0.047 \\
\addlinespace[0.3em] \cmidrule{2-11} \addlinespace[0.3em]
\multirow{3}{*}{\shortstack[l]{\textit{Multi-Generation} \\ \scriptsize (Acc: 70.7 / 82.9)}} & MC-SE & \textbf{0.096} & \textbf{0.621} & \textbf{0.383} & 0.065 & 0.445 & 0.194 & -0.032 & -0.176 & -0.189 \\
 & MC-NSE & 0.103 & 0.572 & 0.329 & 0.067 & 0.403 & 0.168 & -0.036 & -0.169 & -0.161 \\
 & Consistency & 0.126 & 0.287 & 0.285 & \textbf{0.046} & \textbf{0.501} & \textbf{0.480} & -0.080 & 0.214 & 0.195 \\
\bottomrule
\end{tabular}
}
\end{table}

\FloatBarrier

\clearpage
\section{Additional Figures for UQ Shifts}
\label{appendix:additional_figures_UQ_shifts}

\begin{figure}[!ht]
    \centering
    
    \begin{subfigure}{0.65\linewidth}
        \centering
        \includegraphics[width=\linewidth]{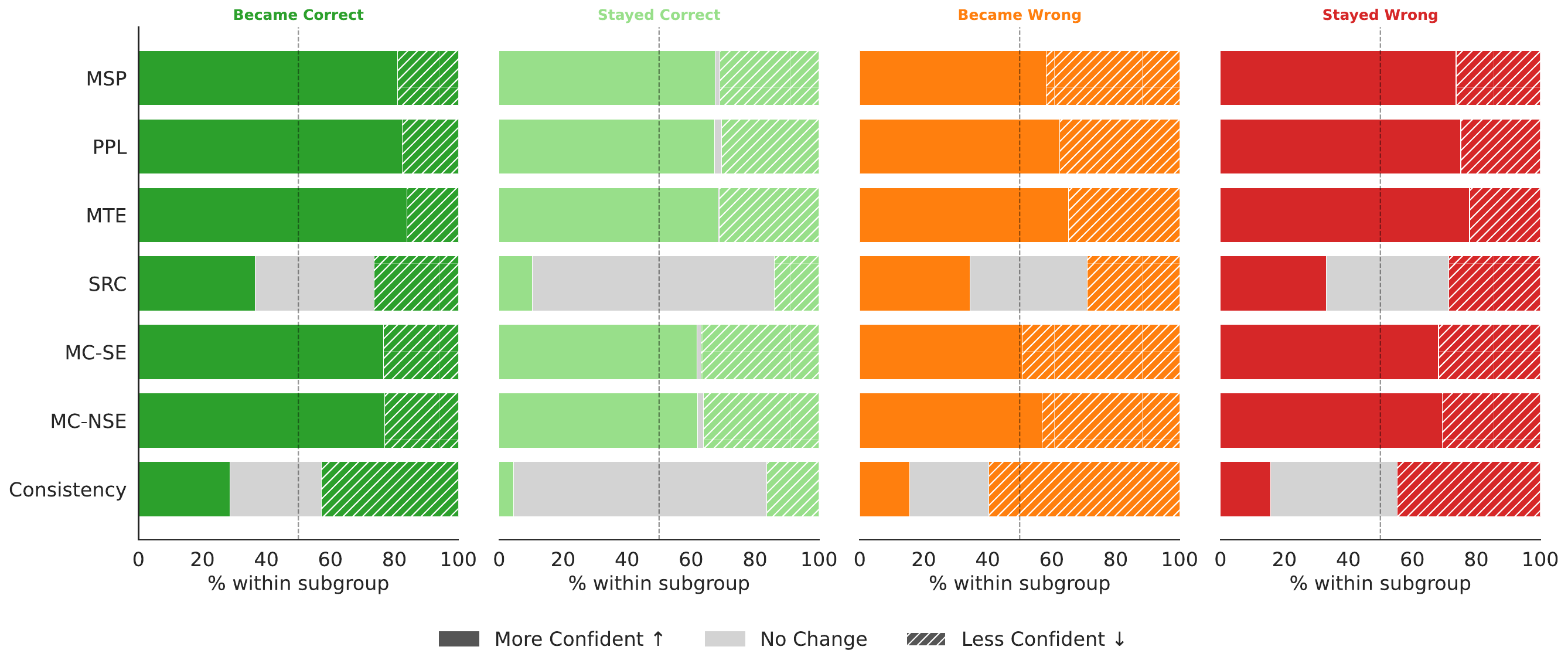}
        \caption{OK-VQA}
        \label{fig:appendix_uq_shift_okvqa}
    \end{subfigure}

    \begin{subfigure}{0.65\linewidth}
        \centering
        \includegraphics[width=\linewidth]{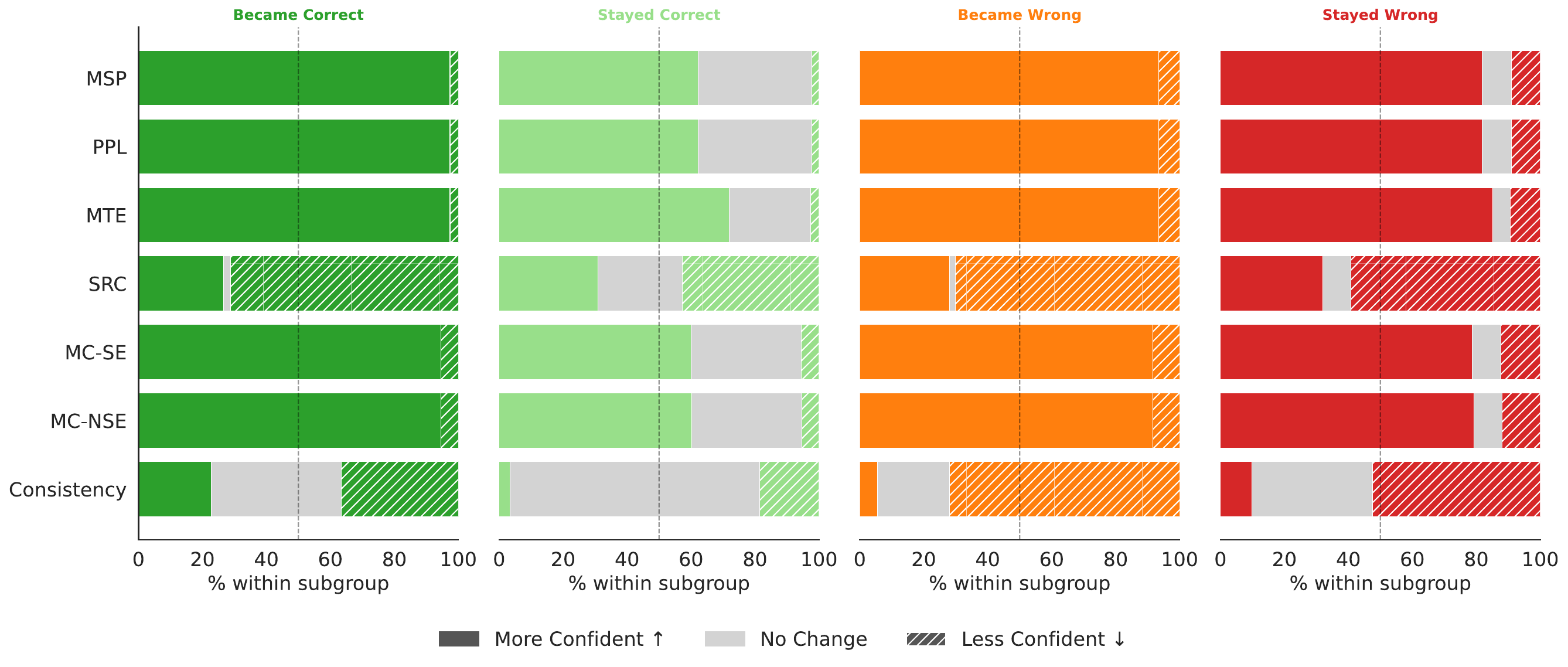}
        \caption{MMMU-Pro-Vision}
        \label{fig:appendix_uq_shift_mmmu}
    \end{subfigure}

    \begin{subfigure}{0.65\linewidth}
        \centering
        \includegraphics[width=\linewidth]{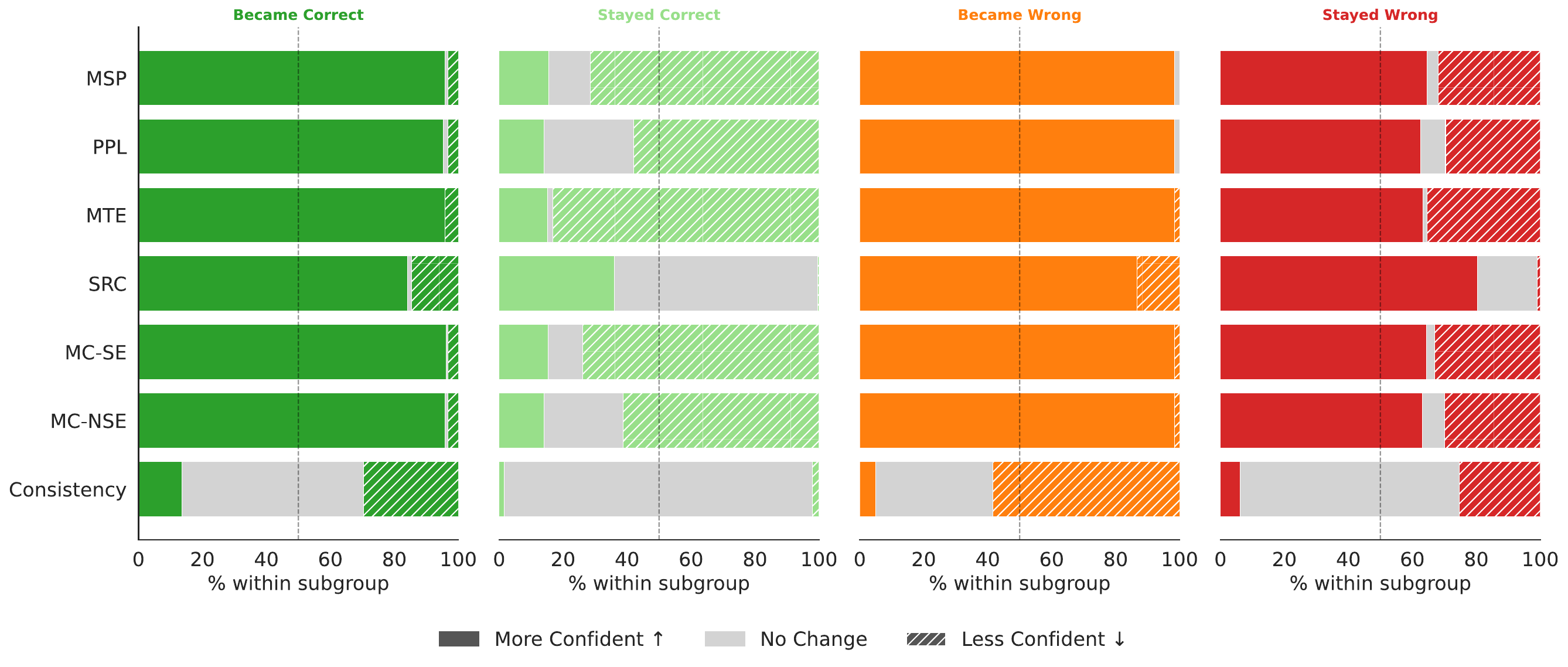}
        \caption{Oxford Pets}
        \label{fig:appendix_uq_shift_oxfordpets}
    \end{subfigure}

    \caption{Sample-level confidence shifts under Chain-of-Thought prompting for Qwen3-VL-32B-Instruct across different datasets. Each subfigure corresponds to a dataset and reports confidence changes conditioned on correctness transitions.}
    \label{fig:appendix_uq_shift_all}
\end{figure}

\FloatBarrier
\clearpage
\section{Answer and Reasoning lengths}
\label{appendix:token_lengths}

\begin{table}[!ht]
\centering
\small
\setlength{\tabcolsep}{4pt}
\renewcommand{\arraystretch}{1.1}
\caption{Average answer and reasoning lengths (number of tokens; mean $\pm$ std).}
\label{tab:lengths}
\resizebox{\linewidth}{!}{%
\begin{tabular}{llccc}
\toprule
 &  & \multicolumn{1}{c}{Answer Tokens} & \multicolumn{1}{c}{Answer Tokens} & \multicolumn{1}{c}{Reasoning Tokens} \\
\cmidrule(lr){3-3}\cmidrule(lr){4-4}\cmidrule(lr){5-5}
Model & Dataset & \multicolumn{1}{c}{No Reasoning} & \multicolumn{1}{c}{Reasoning} & \multicolumn{1}{c}{Reasoning} \\
\midrule
\multirow{4}{*}{Qwen3-VL-32B-IT} & MMMU Pro Vision & 1.0 $\pm$ 0.0 & 1.0 $\pm$ 0.0 & 521.9 $\pm$ 1742.1 \\
 & MathVista & 1.8 $\pm$ 1.8 & 1.4 $\pm$ 1.1 & 206.8 $\pm$ 842.9 \\
 & OK-VQA & 2.0 $\pm$ 1.0 & 2.2 $\pm$ 1.2 & 47.8 $\pm$ 24.7 \\
 & Oxford Pets & 1.7 $\pm$ 0.4 & 1.7 $\pm$ 0.4 & 60.9 $\pm$ 11.3 \\
\addlinespace[0.35em]
\multirow{4}{*}{Qwen3-VL-8B-IT} & MMMU Pro Vision & 1.0 $\pm$ 0.0 & 1.0 $\pm$ 0.0 & 186.0 $\pm$ 439.2 \\
 & MathVista & 1.5 $\pm$ 1.5 & 1.6 $\pm$ 1.6 & 168.3 $\pm$ 595.0 \\
 & OK-VQA & 2.0 $\pm$ 0.9 & 2.2 $\pm$ 1.1 & 48.6 $\pm$ 16.0 \\
 & Oxford Pets & 3.8 $\pm$ 0.5 & 3.8 $\pm$ 0.4 & 164.0 $\pm$ 88.7 \\
\addlinespace[0.35em]
\multirow{4}{*}{Qwen3-VL-8B-Thinking} & MMMU Pro Vision & -- & 1.0 $\pm$ 0.0 & 1879.1 $\pm$ 1952.3 \\
 & MathVista & -- & 1.4 $\pm$ 1.3 & 1373.4 $\pm$ 2317.1 \\
 & OK-VQA & -- & 2.0 $\pm$ 1.0 & 294.7 $\pm$ 407.8 \\
 & Oxford Pets & -- & 1.8 $\pm$ 0.4 & 127.5 $\pm$ 213.6 \\
\addlinespace[0.35em]
\multirow{4}{*}{Gemma3-4B-IT} & MMMU Pro Vision & 1.0 $\pm$ 0.2 & 1.0 $\pm$ 0.3 & 89.3 $\pm$ 45.8 \\
 & MathVista & 2.2 $\pm$ 2.2 & 1.8 $\pm$ 2.2 & 103.4 $\pm$ 110.7 \\
 & OK-VQA & 1.8 $\pm$ 0.9 & 2.0 $\pm$ 1.0 & 40.7 $\pm$ 7.9 \\
 & Oxford Pets & 1.8 $\pm$ 0.4 & 1.8 $\pm$ 0.4 & 46.2 $\pm$ 7.6 \\
\bottomrule
\end{tabular}
}
\end{table}

\section{Confidence Progression Throughout the Reasoning Trace}
\label{appendix:confidence_progression}

\begin{figure}[t]
    \centering
    \includegraphics[width=0.5\linewidth]{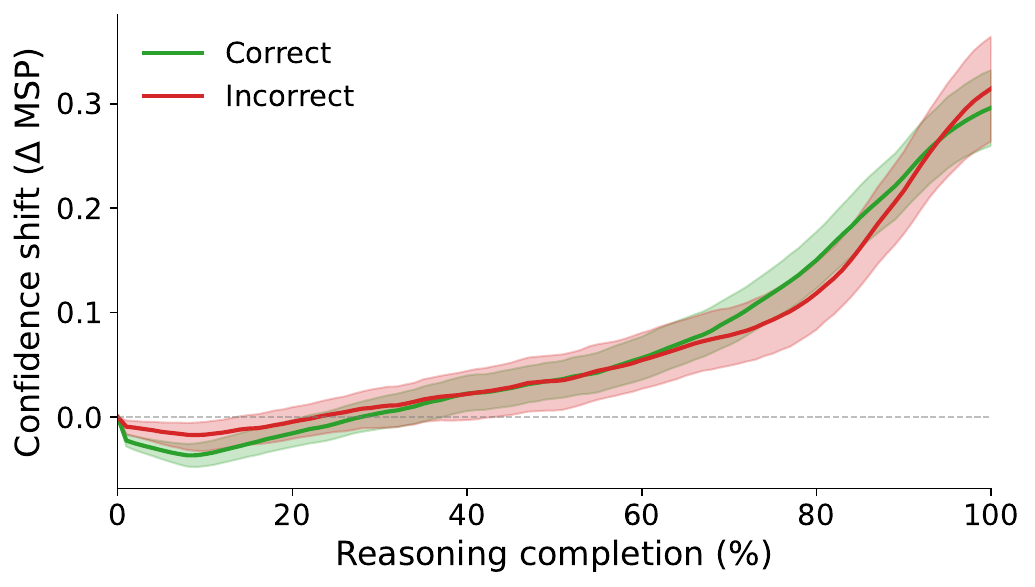}
   \caption{Confidence shift in the final predicted answer's MSP as a function of reasoning progress. Curves are split by whether the final answer was correct or incorrect. Solid lines show the mean, and shaded regions show the 95\% confidence interval estimated via bootstrapping.}
    \label{fig:confidence_progression}
\end{figure}

We evaluate how the final answer's MSP evolves over the course of a reasoning trace on MathVista, using Qwen3-VL-8B-IT with CoT prompting. At each reasoning token position $t$, we condition the model on the reasoning prefix $r_{\leq t}$ followed by the \texttt{<answer>} tokens, and compute the likelihood of the model's final predicted answer $\hat{a}$ under this conditioning. We then subtract the baseline likelihood computed with no reasoning tokens generated:
\begin{equation}
    \Delta\text{MSP}(t) = P(\hat{a} \mid x, r_{\leq t}, \theta) - P(\hat{a} \mid x, \theta),
\end{equation}
where $x$ is the input. Figure~\ref{fig:confidence_progression} shows this confidence shift separately for traces that ultimately produce a correct versus incorrect final answer. The two curves track each other closely throughout reasoning, indicating that the model's confidence in its eventual answer grows at a similar rate regardless of whether that answer will turn out to be correct.

\FloatBarrier
\section{Answer Frequency Diagrams}
\label{appendix:answer_freq_aditional_figures}

\begin{figure}[!ht]
     \centering
     \begin{subfigure}[t]{0.48\textwidth}
         \centering
         \includegraphics[width=\textwidth]{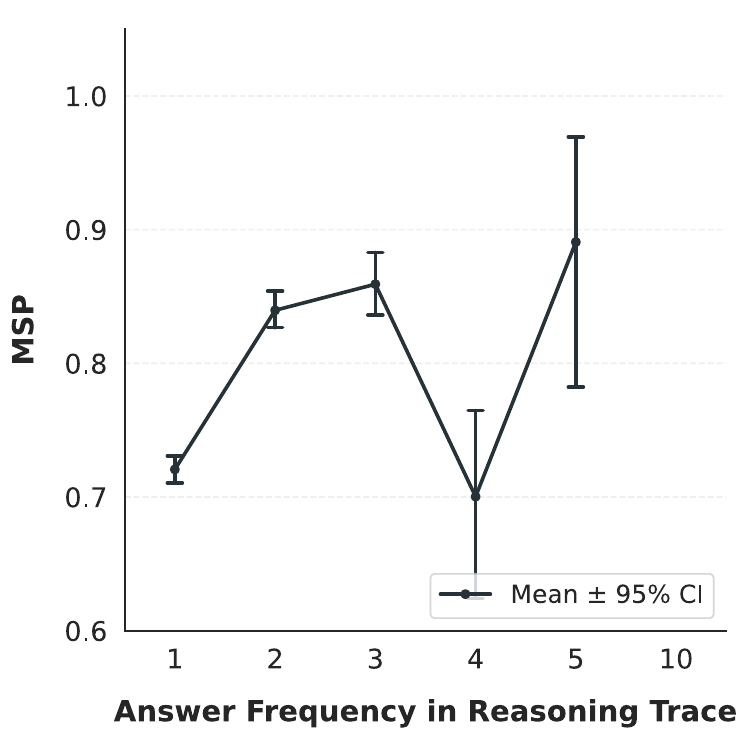}
         \caption{Gemma3-4b-IT}
         \label{fig:freq_hist_gemma3}
     \end{subfigure}
     \hfill
     \begin{subfigure}[t]{0.48\textwidth}
         \centering
         \includegraphics[width=\textwidth]{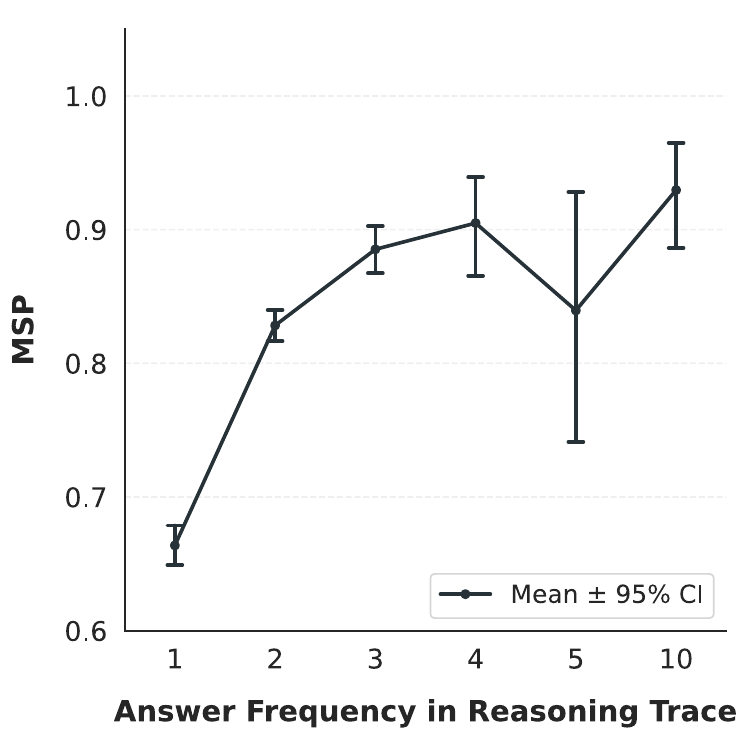}
         \caption{Qwen3-VL-8b-IT}
         \label{fig:freq_hist_qwen_8b}
     \end{subfigure}
     \hfill
     \caption{\textbf{Reasoning Trace Analysis on Incorrect Samples.} On OK-VQA, reasoning traces that mention the predicted answer more frequently tend to yield higher final-answer MSP, even when the prediction is incorrect.}
     \label{fig:appendix_answer_freq_figure}
\end{figure}

\FloatBarrier
\section{Repeated Reasoning and Answering.}
\label{appendix:toy_experiment}

\begin{figure}[tb]
    \centering
    \includegraphics[width=0.5\textwidth]{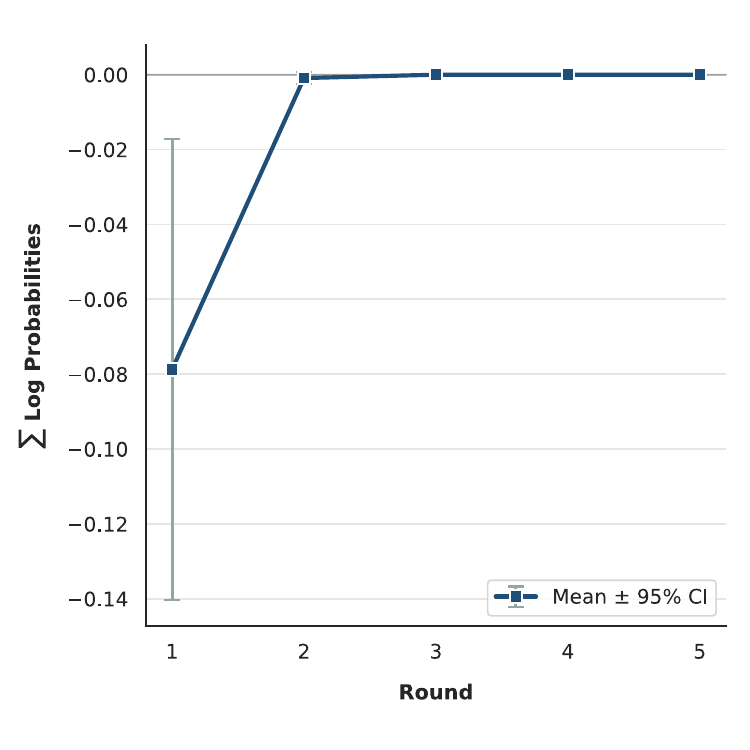}
    \caption{Sequential Confidence Trajectory. The log-sum-probability of the answer tokens across multiple rounds of reasoning and answering. Qwen3-VL-8B-IT on a random subset (n=200) of MathVista.}
    \label{fig:toy-experiment}
\end{figure}

To better understand how the final answer conditions on past possible answers present in the reasoning trace, we designed a controlled toy experiment. We prompted Qwen3-VL-8B-IT to reason and answer in the standard CoT prompt format, then extended the conversation over multiple turns, asking the model to reason further and answer again while all previous reasoning traces and answers remained in context. We ran this on a random subset of MathVista (n=200) and tracked the sum of the log-probabilities over the answer tokens for each instance of answering. As shown in \cref{fig:toy-experiment}, the first-round answer carries a meaningful (non-zero) log-probability. After just one round of conditioning on a prior answer and reasoning trace, subsequent log-probabilities collapse toward zero, indicating near-perfect apparent confidence. This ``confidence collapse'' is not earned: the model is not becoming more accurate across rounds, it is merely becoming more committed to repeating itself.

This toy experiment illustrates implicit answer conditioning in its most extreme form. Even in standard single-turn CoT prompting, a reasoning trace that builds strongly toward a conclusion creates the same conditioning pressure on the final answer tokens, just to a lesser degree. Crucially, the answer need not appear verbatim in the reasoning trace for this effect to occur, it is sufficient for the reasoning to converge semantically on a conclusion.


\FloatBarrier
\newpage
\section{Answer Masking in Reasoning Traces: MathVista}
\label{appendix:masked_cot_mathvista}
\begin{table}[!ht]
\centering
\caption{\textbf{Masked-CoT Results on MathVista using Qwen3-VL-8B-IT.} Masking answer mentions partially restores uncertainty ranking, while random masking of an equivalent number of reasoning tokens degrades ranking quality. These results are computed solely on the subset of image/question pairs where the reasoning traces include the predicted final answer.}
\label{tab:masked_mathvista}
\scriptsize
\setlength{\tabcolsep}{4pt}
\renewcommand{\arraystretch}{1.05}
\resizebox{\linewidth}{!}{
\begin{tabular}{l ccc ccc ccc ccc}
\toprule
 & \multicolumn{3}{c}{No-CoT (Acc 62.9)}
 & \multicolumn{3}{c}{CoT (Acc 72.9)}
 & \multicolumn{3}{c}{Masked-CoT (Acc 72.9)}
 & \multicolumn{3}{c}{Random Mask (Acc 72.9)} \\
\cmidrule(lr){2-4} \cmidrule(lr){5-7} \cmidrule(lr){8-10} \cmidrule(lr){11-13}
UQ Method & AUGRC$\downarrow$ & Spear.$\uparrow$ & PRR$\uparrow$
& AUGRC$\downarrow$ & Spear.$\uparrow$ & PRR$\uparrow$
& AUGRC$\downarrow$ & Spear.$\uparrow$ & PRR$\uparrow$
& AUGRC$\downarrow$ & Spear.$\uparrow$ & PRR$\uparrow$ \\
\midrule
MSP & 0.142 & 0.313 & 0.522 
    & 0.130 & 0.043 & 0.115 
    & \textbf{0.129} & \textbf{0.055} & \textbf{0.200}
    & 0.147 & -0.083 & -0.032 \\

MTE & 0.143 & 0.304 & 0.526 
    & 0.134 & 0.013 & 0.075 
    & \textbf{0.119} & \textbf{0.133} & \textbf{0.264}
    & 0.144 & -0.062 & 0.010 \\

PPL & 0.145 & 0.289 & 0.513 
    & 0.134 & 0.017 & 0.100 
    & \textbf{0.131} & \textbf{0.038} & \textbf{0.184}
    & 0.148 & -0.094 & -0.050 \\
\bottomrule
\end{tabular}
}
\end{table}

\end{document}